\documentclass{article}

\usepackage[preprint]{neurips_2026}
\makeatletter
\renewcommand{\@notice}{}
\makeatother

\usepackage[utf8]{inputenc} 
\usepackage[T1]{fontenc}    
\usepackage{hyperref}       
\usepackage{url}            
\usepackage{booktabs}       
\usepackage{amsmath}        
\usepackage{amsfonts}       
\usepackage{amssymb}        
\usepackage{bm}             
\usepackage{nicefrac}       
\usepackage{microtype}      
\usepackage[dvipsnames]{xcolor}  
\usepackage{caption}        
\usepackage{graphicx}       
\usepackage{xspace}
\usepackage{wrapfig}        
\usepackage{seqsplit}
\usepackage{multirow}
\usepackage{longtable}
\usepackage{tabularx}
\usepackage{array}

\newcommand{\PPG}{PP-GNN\xspace}
\newcommand{\PPGs}{PP-GNNs\xspace}
\newcommand{\MPG}{MP-GNN\xspace}
\newcommand{\MPGs}{MP-GNNs\xspace}
\newcommand{\modelname}{FilterMoE\xspace}
\newcommand{\MoE}{MoE}
\title{Gate the Filter, Not the Message: Node-Channel Mixtures for Pre-Propagation GNNs}

\author{%
  Zichao Yue$^{1}$\thanks{Correspondence to: Zichao Yue \texttt{<zy383@cornell.edu>}.} \quad Zhiru Zhang$^{1}$\\[0.5em]
  $^{1}$School of Electrical and Computer Engineering, Cornell University\\
  Ithaca, New York, USA
}

\date{}

\begin{document}

\maketitle

\begin{abstract}
Pre-propagation graph neural networks (\PPGs) push all graph-dependent computation into a preprocessing step and train only on the resulting dense hop features, which makes them highly scalable. A puzzle in this regime is that more complex hop aggregators do not reliably outperform simpler ones: on many benchmarks, a plain MLP-based aggregator matches or beats hop-attention variants. We revisit this behavior from a graph-filter perspective. Over a precomputed diffusion basis, existing \PPGs differ mainly in how filter coefficients are shared across nodes and feature channels, rather than simply in raw aggregator capacity. MLP-based architectures learn channel-dependent filters that are largely shared across nodes, while hop-attention-based architectures learn node-dependent mixtures that are largely shared across channels. This reveals a missing regime in standard \PPG designs: joint node- and channel-adaptive filtering under the pre-propagation computational contract. We propose \modelname{}, a mixture-of-experts \PPG in which a small bank of learnable Chebyshev filter experts is routed jointly over nodes and channels by a 3D gating tensor. 
Across eleven homophilic and heterophilic benchmarks, \modelname{} outperforms strong \PPG baselines on nine datasets and ranks first on all three large-scale benchmarks, improving the average test score by 1.53 points. These results establish joint node–channel filter routing as a robust alternative to dataset-specific hop-aggregator selection.
\end{abstract}


\section{Introduction}

Graph neural networks (GNNs) have become a standard tool for learning on relational data, but their scalability is often limited by repeated neighborhood aggregation during training \citep{gilmer2017neural,hamilton2017sage}. In message-passing-based GNNs (\MPGs), aggregation is performed at every layer and every epoch, leading to irregular sparse computation, high memory traffic, and the well-known neighbor-explosion problem \citep{hamilton2017sage}. Pre-propagation GNNs (\PPGs) address this bottleneck by moving message passing to a preprocessing stage and training only on the resulting dense node features \citep{frasca2020sign,zhang2022gamlp,deng2024hoga}. This decoupled design makes \PPGs attractive for large-scale node classification, combining one-time graph preprocessing with up to two orders of magnitude faster training on large graph benchmarks \citep{frasca2020sign,zhang2022gamlp,yue2025graph}.

\begin{wrapfigure}{r}{0.6\linewidth}
  \vspace{-10pt}
  \centering
  \includegraphics[width=\linewidth]{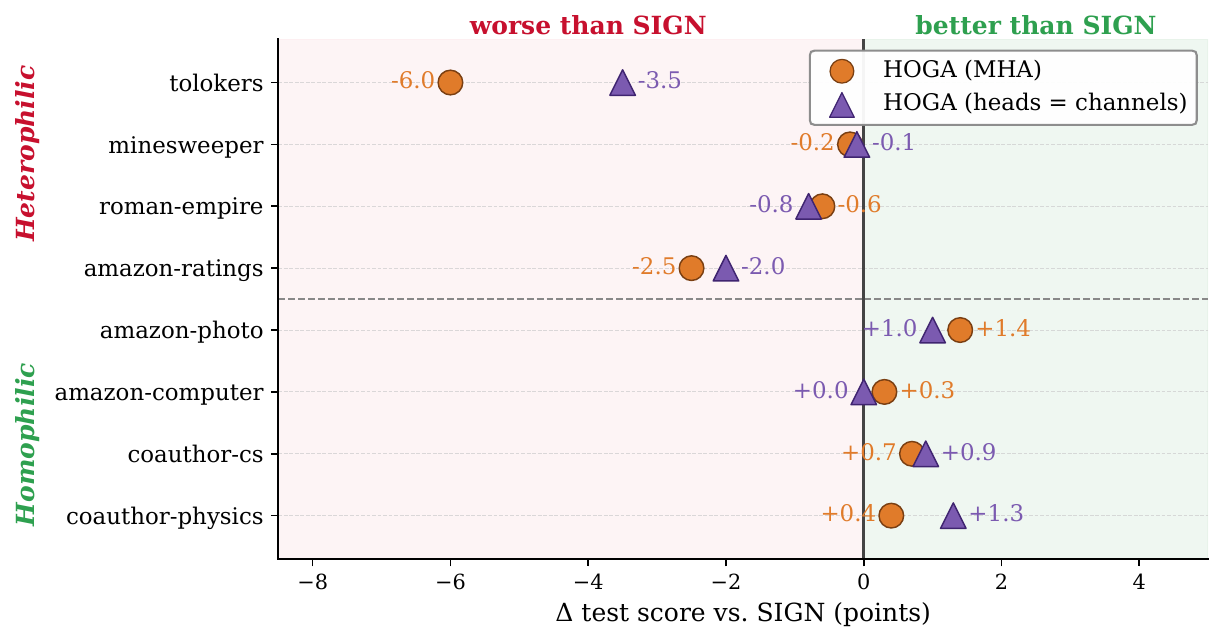}
  \caption{$\Delta$ test score (\%) of HOGA vs.\ SIGN on eight graphs. HOGA loses to SIGN on heterophilic graphs while outperforming SIGN on homophilic ones. Setting HOGA $\mathrm{heads}=\mathrm{hidden\_channels}$ (triangles) does not close the gap consistently.}
  \label{fig:motivation}
  \vspace{-10pt}
\end{wrapfigure}

Given the preprocessed node features, \PPGs differ mainly in the mechanisms employed for feature aggregation. However, a more complex feature aggregator does not necessarily lead to better accuracy. One might expect an attention-based feature aggregator, such as the one used in HOGA, to dominate a simpler MLP-based model such as SIGN. Our experiments do not support such a uniform conclusion: on several datasets, SIGN is as strong as or stronger than HOGA, as shown in Figure~\ref{fig:motivation}. This observation suggests that the main limitation of current \PPGs is not feature-aggregator capacity alone.

A useful way to explain this behavior is through the lens of graph filtering. Let
\[
\mathbf{Z}=\left[\mathbf{X},\mathbf{S}\mathbf{X},\mathbf{S}^2\mathbf{X},\ldots,\mathbf{S}^K\mathbf{X}\right]
\]
denote the stack of precomputed node features generated by a diffusion operator $\mathbf{S}$. A \PPG then learns a predictor over this diffusion basis. Assuming a simple linear aggregator, this view becomes
\[
f_\theta(\mathbf{Z})=\sum_{k=0}^{K}\mathbf{S}^k\mathbf{X}\mathbf{W}_k,
\]
which can be interpreted as learning a graph filter over the precomputed diffusion basis.
When the aggregator is more complex, as in most \PPGs, the learned object can be viewed as an implicit filter over the same precomputed basis.
From this perspective, the key distinction between existing \PPGs is not raw architectural complexity, but \emph{how coefficients over the diffusion basis are shared}. SIGN realizes a channel-dependent but largely node-shared filter, whereas hop-wise attention variants such as HOGA induce node-dependent but largely channel-shared mixtures. Thus, a more expressive hop aggregator is not automatically better: it may provide a different form of adaptivity, rather than a universally stronger one.

This view immediately raises a second question: why not simply combine both forms of adaptivity by increasing the number of attention heads or by directly predicting nearly independent filters for all node--channel pairs? Empirically, we find that this strategy is ineffective. In particular, setting the number of HOGA heads equal to the channel dimension does not yield consistent gains, as shown in Figure~\ref{fig:motivation}. Conceptually, such a design imposes a weak inductive bias: instead of assuming that many node--channel pairs share a compact repertoire of useful spectral responses, it effectively asks the model to generate almost unconstrained filters for a very large number of pairs. Prior work in spectral GNNs has also largely explored channel-adaptive filtering and node-adaptive filtering in separate lines \citep{zhao2021adaptive,dong2021adagnn,guo2023graph}, leaving their joint realization largely unexplored.

We therefore propose \modelname{}, a mixture-of-experts (MoE) \PPG that introduces joint node- and channel-adaptive filtering while preserving the pre-propagation paradigm. Our key idea is to replace direct prediction of node--channel-specific polynomial coefficients with a small bank of learnable Chebyshev filter experts and an input-dependent routing mechanism. Concretely, each expert is a Chebyshev polynomial filter, which provides an explicit spectral response and efficient preprocessing.
A 3D gating tensor $\mathbf{G}\in\mathbb{R}^{N\times M\times F}$ then mixes the $M$ experts jointly across nodes and feature channels, so that each node--channel pair receives its own effective filter while all filters remain compositions of a small shared bank. We instantiate the router in two practical forms: a simple MLP router directly mapping the expert outputs to the 3D gating tensor, and an attention-style router that matches node representations to learned expert embeddings and then applies channel-wise adjustment. To prevent the expert bank from collapsing to near-duplicate filters, we further introduce a diversity regularizer defined in spectral-response space rather than coefficient space.

Across 11 benchmark graphs, \modelname{} provides a robust alternative to manually selecting a hop aggregator for each dataset. Compared with the strongest \PPG baseline on each benchmark, \modelname{} improves the average test score by \textbf{1.53} points, and wins on \textbf{9 of 11} datasets. The advantage is especially clear in the large-graph regime where \PPGs are most useful: \modelname{} ranks first on all three large-scale benchmarks, outperforming the best \PPG baseline by \textbf{+1.76} on pokec, \textbf{+0.93} on ogbn-products, and \textbf{+0.71} on ogbn-papers100M. Compared with \MPGs, \modelname{} achieves competitive accuracy while preserving the pre-propagation training contract. On the three large graphs, it outperforms all the evaluated \MPG baselines. Overall, these results suggest that the main limitation of existing \PPGs is not the lack of a more complex hop-feature aggregator, but the lack of a structured mechanism for learning graph filters that are jointly node- and channel-adaptive while retaining the scalability advantage of pre-propagation.

\paragraph{Contributions.}
Our main contributions are as follows:
\begin{itemize}
    \item We use a graph-filter perspective to analyze existing \PPG hop aggregators and show that their main difference lies in how coefficients over a precomputed diffusion basis are shared across nodes and channels, which explains why more complex hop aggregators do not necessarily outperform simpler ones.
    \item We identify the missing regime of joint node- and channel-adaptive filtering in \PPGs, and propose \modelname{}, a mixture-of-experts \PPG that routes a small bank of shared Chebyshev filter experts to each node--channel pair through a 3D gating mechanism.
    \item We show that \modelname{} acts as a robust learned alternative to manual hop-aggregator selection: it improves the average test score over the strongest \PPG baselines on each dataset by 1.53 points, winning on 9 of 11 benchmarks.
    \item We conduct extensive experiments and ablations to quantify the effects of node adaptivity, channel adaptivity, routing design, diffusion-operator effects, and auxiliary regularization across a broad collection of graphs.
\end{itemize}

\section{Background and Related Work}
\label{sec:bg_related_work}
\label{sec:related_work}

\paragraph{Pre-propagation GNNs.}
Message-passing GNNs (\MPGs) repeatedly aggregate over graph neighborhoods during training, which couples model depth to neighborhood expansion and makes large-graph mini-batching challenging~\citep{hamilton2017sage}. Pre-propagation GNNs (\PPGs{}) decouple these two steps: graph-dependent propagation is performed once, and training uses only cached dense node features. For a single diffusion operator $\mathbf{S}\in\mathbb{R}^{N\times N}$ derived from the graph, such as a normalized adjacency or random-walk-based operator, a standard \PPG{} constructs
\begin{equation}
\mathbf{Z}
=
\big[
\mathbf{X}\,\Vert\,\mathbf{S}\mathbf{X}\,\Vert\,\mathbf{S}^{2}\mathbf{X}\,\Vert\,\cdots\,\Vert\,\mathbf{S}^{K}\mathbf{X}
\big]
\in\mathbb{R}^{N\times (K+1)F},
\label{eq:bg_ppg_stack}
\end{equation}  where $\mathbf{X}\in\mathbb{R}^{N\times F}$ is the node-feature matrix, and $K$ is the number of hops.
Then a row-wise dense predictor $\hat{\mathbf y}_i=f_\theta(\mathbf z_i)$ is trained for the downstream task without further sparse message passing. This preprocessing-only contract underlies the scalability of SGC and SIGN~\citep{wu2019sgc,frasca2020sign}; later \PPGs{} mainly differ in how they fuse the cached hop features. SAGN uses hop attention with self-label-enhanced training~\citep{sun2025sagnsle}, GAMLP learns adaptive fusion over multi-hop feature and label propagations~\citep{zhang2022gamlp}, and HOGA applies gated self-attention to hop-wise features~\citep{deng2024hoga}. Recent systems work further analyzes the input-expansion and data-loading bottlenecks of this family~\citep{yue2025graph}. Appendix~\ref{app:extended_bg_related_work} gives a more detailed PP-GNN formalization and representative model equations.

\paragraph{Adaptive graph filtering.}
Spectral GNNs view propagation as graph filtering, with polynomial filters providing a practical bridge between spectral responses and localized graph computation~\citep{defferrard2016convolutional,he2022chebnetii}. Many models relax the assumption that one global filter should be shared everywhere. On the channel side, diffusion and spectral models such as DCNN, ADC, AdaGNN, JacobiConv, and $\omega$GNN learn multiple diffusion scales, per-channel diffusion radii, adaptive hop weights, polynomial coefficients, or mixtures of propagation operators~\citep{atwood2016diffusion,zhao2021adaptive,dong2021adagnn,wang2022powerful,eliasof2023improving}. These methods increase flexibility across feature dimensions, but their filters are still largely shared across nodes. On the node side, spatial models such as GAT, ACM-GCN, and AGDN introduce node-, channel-, or hop-dependent aggregation weights~\citep{velickovic2017gat,luan2022revisiting,sun2020adaptive}, while node-oriented spectral methods such as NFGNN, DSF, and node-variant graph filters make filtering behavior itself node dependent~\citep{zheng2023node,guo2023graph,gama2021nodevariant}. In contrast, \modelname{} realizes joint node--channel adaptivity in the \PPG{} regime by routing over a shared learnable Chebyshev expert bank, rather than predicting nearly independent coefficients for every node--channel pair.

\paragraph{Mixture-of-experts on graphs.}
Mixture-of-experts (\MoE{}) models provide conditional computation through input-dependent routing~\citep{jacobs1991adaptive}. Sparse \MoE{} layers scale this idea by activating only a subset of experts per input, as in sparsely gated \MoE{}, GShard, and Switch Transformer~\citep{shazeer2017outrageously,lepikhin2020gshard,fedus2022switch}. Graph learning has recently adopted \MoE{} routing in several forms. GraphDIVE~\citep{park2024graphdive} applies \MoE{} to graph classification under class imbalance. GMoE learns node-wise selection among neighborhood aggregation experts for node classification \citep{wang2023gmoe}. Mowst combines a weak MLP expert with a stronger GNN expert using confidence-guided routing~\citep{park2023mowst}. Node-MoE uses \MoE{} for node-wise filter selection, and GNNMoE combines decoupled message-passing experts with soft or hard routing~\citep{chen2024nodemoe,wang2024gnnmoe}. These models demonstrate the value of routing on graphs, but most route among encoders, message-passing backbones, or node-wise experts. \modelname{} instead applies \MoE{} directly to \emph{pre-propagation graph filters}: the experts are spectral responses over a cached basis, and the router mixes them jointly over nodes and feature channels while preserving dense \PPG{} training.

\section{Method}
\label{sec:method}

\subsection{Overview and Notation}
\label{sec:method-overview}

\begin{figure}[t]
  \centering
  \includegraphics[width=\linewidth]{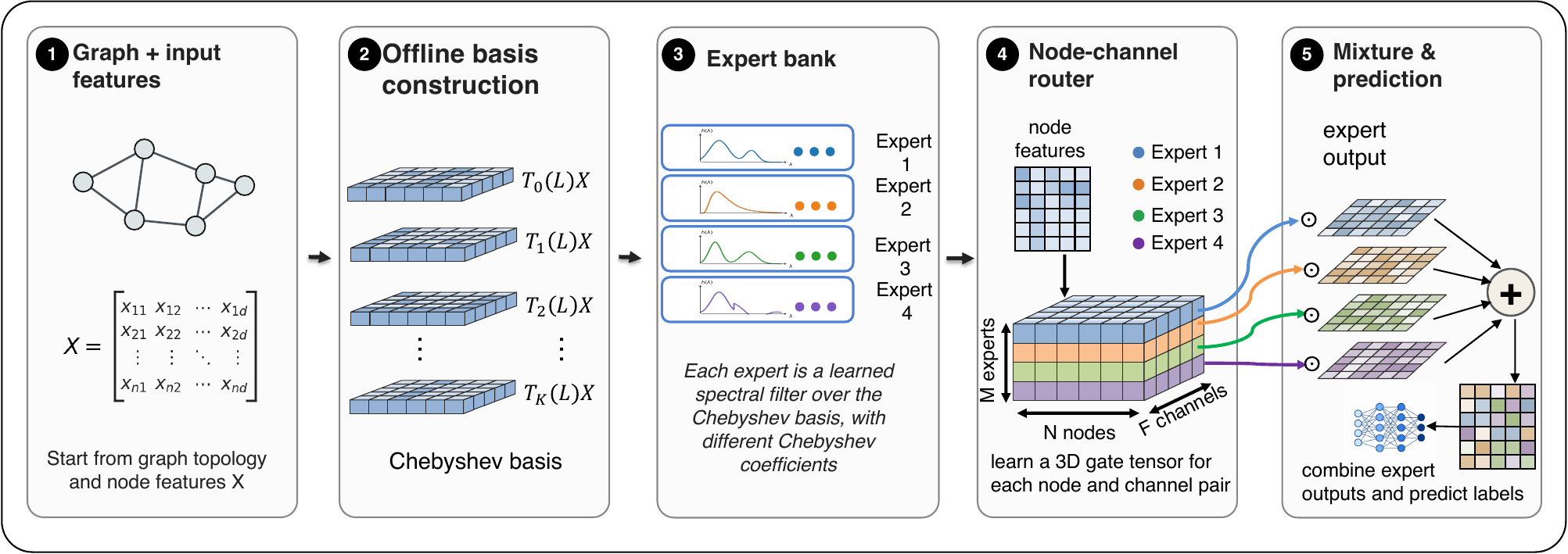}
  \caption{Architecture overview of \modelname}
  \label{fig:schema}
  \vspace{-5pt}
\end{figure}
Let $\mathbf{X}\in\mathbb{R}^{N\times F}$ be the node-feature matrix and let $\tilde{\mathbf{L}}$ denote the rescaled graph Laplacian with spectrum in $[-1,1]$. \modelname{} follows the \PPG{} contract: graph-dependent propagation is performed once before training. We materialize a Chebyshev diffusion basis
\begin{equation}
  \mathbf{B}_{k}=T_{k}(\tilde{\mathbf{L}})\mathbf{X},
  \qquad k=0,\ldots,K,
  \label{eq:filtermoe-cheb-basis}
\end{equation}
where $T_k(\cdot)$ is the $k$-th Chebyshev polynomial and $\mathbf{B}_{k}\in\mathbb{R}^{N\times F}$.

\modelname{} learns a small bank of $M$ Chebyshev filter experts and routes each node--channel pair over this shared bank. This gives an effective filter $\boldsymbol{\beta}_{i,f}$ that can vary with node $i$ and feature channel $f$, while avoiding an unconstrained filter for every node--channel pair. The routed representation is passed to a dense predictor, so no sparse message passing is performed during training. The overall architecture is shown in Fig~\ref{fig:schema}.

\subsection{Chebyshev Filter Experts}
\label{sec:method-experts}

Expert $m$ is parameterized by Chebyshev coefficients $\boldsymbol{\alpha}_{m}\in\mathbb{R}^{K+1}$ and realizes the spectral response
\begin{equation}
  g_{m}(\lambda)=\sum_{k=0}^{K}\alpha_{m,k}T_{k}(\lambda).
  \label{eq:filtermoe-expert-response}
\end{equation}
Because the basis in Eq.~\eqref{eq:filtermoe-cheb-basis} has already been materialized, applying an expert during training requires only a dense combination of cached tensors:
\begin{equation}
  \mathbf{H}^{(m)}=
  \sum_{k=0}^{K}\alpha_{m,k}\mathbf{B}_{k}
  \in\mathbb{R}^{N\times F}.
  \label{eq:filtermoe-expert-output}
\end{equation}
Thus, \modelname{} does not reintroduce per-epoch message passing or sparse neighborhood aggregation. In practice, we initialize the expert coefficients using the interpolation-based Chebyshev parameterization of ChebNetII~\citep{he2022chebnetii}, which provides stable initial polynomial responses, and then learn the coefficients during training.

\paragraph{Spectral response sketches.}
The response-aware router and the diversity loss both need a compact way to
reason about what each learned filter does on the current graph. We use stochastic
Lanczos quadrature (SLQ) once in preprocessing to form a small graph-specific
spectral grid~\citep{ubaru2017fast}:
\begin{equation}
  \{(\theta_p,w_p)\}_{p=1}^{P},
  \qquad \theta_p\in[-1,1],\quad w_p\ge 0,
  \quad \sum_{p=1}^{P}w_p=1.
  \label{eq:filtermoe-slq-grid}
\end{equation}
For a spectral function $a(\lambda)$, this grid approximates spectral integration by
\begin{equation}
  \int a(\lambda)\,d\mu_{\tilde{\mathbf{L}}}(\lambda)
  \approx
  \sum_{p=1}^{P}{w}_p a(\theta_p).
  \label{eq:filtermoe-slq-average}
\end{equation}
We evaluate each expert response on this grid using the fixed Chebyshev Vandermonde matrix
\begin{equation}
  \boldsymbol{\Phi}\in\mathbb{R}^{P\times(K+1)},
  \qquad \Phi_{p,k}=T_{k}(\theta_{p}),
  \label{eq:filtermoe-slq-phi}
\end{equation}
and obtain the sampled response curve
\begin{equation}
  r_{m,p}=g_m(\theta_p),
  \qquad
  \mathbf{r}_{m}=\boldsymbol{\Phi}\boldsymbol{\alpha}_{m}
  \in\mathbb{R}^{P}.
  \label{eq:filtermoe-sampled-response}
\end{equation}
The router uses the weighted response sketch
\begin{equation}
  \widehat{r}_{m,p}=\sqrt{{w}_p}\,r_{m,p},
  \qquad
  \widehat{\mathbf r}_m
  =\left[\widehat r_{m,1},\ldots,\widehat r_{m,P}\right]^{\top}.
  \label{eq:filtermoe-weighted-response-sketch}
\end{equation}
The weighting makes Euclidean comparisons between sketches approximate spectral inner products on the graph, e.g.,
$\langle \widehat{\mathbf r}_m,\widehat{\mathbf r}_n\rangle
=\sum_p w_p g_m(\theta_p)g_n(\theta_p)$.

\subsection{Joint Node--Channel Routing}
\label{sec:method-routing}

The router produces a gate tensor
\begin{equation}
  \mathbf{G}\in\mathbb{R}^{N\times M\times F},
\end{equation}
where $G_{i,m,f}$ is the weight assigned to expert $m$ for node $i$ and feature channel $f$. Given logits $\mathbf{L}\in\mathbb{R}^{N\times M\times F}$, dense routing normalizes over the expert dimension,
\begin{equation}
  G_{i,m,f}
  =
  \frac{\exp(L_{i,m,f}/\tau)}{
  \sum_{m'=1}^{M}\exp(L_{i,m',f}/\tau)},
  \label{eq:filtermoe-dense-router-softmax}
\end{equation}
where $\tau$ is the routing temperature. We also support a sparse variant that masks all but the largest expert scores for each $(i,f)$ before renormalization, following sparse MoE routing practice~\citep{shazeer2017outrageously,fedus2022switch}.

The routed representation is
\begin{equation}
  \widetilde{\mathbf{X}}_{i,f}
  =
  \sum_{m=1}^{M}G_{i,m,f}\,\mathbf{H}^{(m)}_{i,f}.
  \label{eq:filtermoe-routed-representation}
\end{equation}
Equivalently, each node--channel pair receives an effective Chebyshev filter
\begin{equation}
  \boldsymbol{\beta}_{i,f}=
  \sum_{m=1}^{M}G_{i,m,f}\,\boldsymbol{\alpha}_{m}.
  \label{eq:filtermoe-effective-filter}
\end{equation}
This factorization gives node--channel adaptivity while constraining all effective filters to lie in the span of $M$ shared spectral experts. We instantiate the logits with two routers.

\paragraph{Direct joint MLP router.}
The direct router predicts the full $(M,F)$ logit grid for each node. Let $\boldsymbol{\phi}_{i}\in\mathbb{R}^{D_{\mathrm{in}}}$ denote a node descriptor formed from the pre-routing expert outputs $\mathbf{H}$, optionally augmented with random-walk structural encodings~\citep{rampasek2022recipe}. The router applies
\begin{equation}
  \mathbf{L}_{i}
  =
  \operatorname{reshape}\!\left(
  \mathrm{MLP}_{\mathrm{joint}}(\boldsymbol{\phi}_{i})
  \right)
  \in\mathbb{R}^{M\times F}.
  \label{eq:filtermoe-joint-router}
\end{equation}
This router makes no explicit structural assumption about expert--channel interactions; each expert--channel pair has its own output head.

\paragraph{Response-aware two-stage router.}
The second router uses the spectral response sketch of each expert as an expert key. We embed the weighted response sketch from Eq.~\eqref{eq:filtermoe-weighted-response-sketch} as
\begin{equation}
  \mathbf{e}_{m}
  =
  \mathrm{MLP}_{\mathrm{exp}}\!\left(\mathrm{LN}(\widehat{\mathbf r}_{m})\right)
  \in\mathbb{R}^{D}.
  \label{eq:filtermoe-expert-key}
\end{equation}
Node queries and channel queries are extracted from the expert outputs:
\begin{equation}
  \mathbf{u}_{i}=\mathrm{MLP}_{\mathrm{node}}(\boldsymbol{\psi}_{i}),
  \qquad
  \mathbf{v}_{i,f}=\mathrm{MLP}_{\mathrm{chan}}(\boldsymbol{\psi}_{i,f}),
  \label{eq:filtermoe-node-channel-query}
\end{equation}
where $\boldsymbol{\psi}_{i}$ and $\boldsymbol{\psi}_{i,f}$ are node-level and channel-level summaries of the pre-routing expert outputs, respectively.

Routing proceeds in two stages. Stage~1 computes a channel-shared prior over experts,
\begin{equation}
  s^{(1)}_{i,m}=\langle\mathbf{u}_{i},\mathbf{e}_{m}\rangle,
  \qquad
  \pi_{i,m}=\frac{\exp(s^{(1)}_{i,m}/\tau)}{
  \sum_{m'=1}^{M}\exp(s^{(1)}_{i,m'}/\tau)}.
  \label{eq:filtermoe-stage1-router}
\end{equation}
Stage~2 refines that prior for each feature channel using the same expert keys,
\begin{equation}
  \Delta_{i,m,f}=\langle\mathbf{v}_{i,f},\mathbf{e}_{m}\rangle,
  \qquad
  L_{i,m,f}=\log(\pi_{i,m}+\varepsilon)+\Delta_{i,m,f}.
  \label{eq:filtermoe-stage2-router}
\end{equation}
An optional top-$K_{1}$ screening step keeps only the most likely experts from Stage~1 as candidates for Stage~2. Compared with the direct router, this response-aware router factorizes the logits into a node-level prior plus a channel-level residual and explicitly conditions routing on the graph-aware spectral shapes of the experts.

The routed representation $\widetilde{\mathbf{X}}$ is finally passed to a dense predictor $q_{\omega}$ for the downstream task. For node classification, $\mathcal{L}_{\mathrm{task}}$ is the supervised loss on labeled training nodes.

\subsection{Training Objective}
\label{sec:method-training}

The primary training objective is the supervised task loss plus lightweight auxiliary terms. 
The first group of auxiliary terms are filter-related regulators, including a filter diversity loss term and filter smooth loss term.
The filter diversity loss is used to avoid filter collapsing. On the SLQ grid, define the weighted-$\ell_2$-normalized response
\begin{equation}
  \tilde{r}_{m,p}
  =
  \frac{r_{m,p}}{
  \sqrt{\sum_{q=1}^{P}{w}_{q}\,r_{m,q}^{2}+\varepsilon}}
  \label{eq:filtermoe-normalized-response}
\end{equation}
and the weighted Gram matrix
\begin{equation}
  \Gamma_{mn}
  =
  \sum_{p=1}^{P}{w}_{p}\,\tilde{r}_{m,p}\tilde{r}_{n,p}
  \approx
  \int \tilde g_m(\lambda)\tilde g_n(\lambda)\,
  d\mu_{\tilde{\mathbf L}}(\lambda),
  \label{eq:filtermoe-response-gram}
\end{equation}
where $\tilde g_m$ denotes the normalized response corresponding to the samples $\tilde r_{m,p}$.
The diversity loss is
\begin{equation}
  \mathcal{L}_{\mathrm{div}}
  =
  \frac{1}{M(M-1)}\sum_{m\ne n}\Gamma_{mn}^{2}.
  \label{eq:filtermoe-diversity-loss}
\end{equation}
This term measures the pairwise similarity between normalized spectral responses of different expert filters; minimizing it discourages duplicate responses and encourages a more diverse expert bank.

In addition, we use a spectral Sobolev-type smoothness penalty that damps high-degree Chebyshev coefficients~\citep{tadmor1986exponential},
\begin{equation}
  \mathcal{L}_{\mathrm{sm}}
  =
  \frac{1}{MK}\sum_{m=1}^{M}\sum_{k=1}^{K}k^{2}\alpha_{m,k}^{2}.
  \label{eq:filtermoe-smoothness-loss}
\end{equation}
For MoE routing, we include standard auxiliary losses. The importance loss~\citep{shazeer2017outrageously} penalizes per-channel imbalance in expert mass:
\begin{equation}
  I_{m,f}=\sum_{i}G_{i,m,f},
  \qquad
  \mathcal{L}_{\mathrm{imp}}
  =
  \frac{1}{F}\sum_{f=1}^{F}\mathrm{CV}^{2}_{m}\!\big(I_{m,f}\big)
  =
  \frac{1}{F}\sum_{f=1}^{F}\frac{\mathrm{Var}_{m}(I_{m,f})}{\big(\mathbb{E}_{m}[I_{m,f}]\big)^{2}+\varepsilon}.
  \label{eq:filtermoe-importance-loss}
\end{equation}
For sparse routing, the load loss~\citep{shazeer2017outrageously} applies the same per-channel coefficient-of-variation penalty to the number of nodes assigned to each expert at each channel,
\begin{equation}
  \ell_{m,f}=\sum_{i}\mathbf{1}[G_{i,m,f}>0],
  \qquad
  \mathcal{L}_{\mathrm{load}}
  =
  \frac{1}{F}\sum_{f=1}^{F}\mathrm{CV}^{2}_{m}\!\big(\ell_{m,f}\big).
  \label{eq:filtermoe-load-loss}
\end{equation}
We also use a router $z$-loss to stabilize logit magnitudes~\citep{zoph2022stmoe},
\begin{equation}
  \mathcal{L}_{z}
  =
  \mathbb{E}_{i,f}\!\left[
  \left(\log\sum_{m=1}^{M}\exp L_{i,m,f}\right)^{2}
  \right].
  \label{eq:filtermoe-z-loss}
\end{equation}

The final objective is
\begin{equation}
  \mathcal{L}
  =
  \mathcal{L}_{\mathrm{task}}
  +\lambda_{\mathrm{sm}}\mathcal{L}_{\mathrm{sm}}
  +\lambda_{\mathrm{imp}}\mathcal{L}_{\mathrm{imp}}
  +\lambda_{\mathrm{load}}\mathcal{L}_{\mathrm{load}}
  +\lambda_{z}\mathcal{L}_{z}
  +\lambda_{\mathrm{div}}\mathcal{L}_{\mathrm{div}},
  \label{eq:filtermoe-overall-objective}
\end{equation}
where all $\lambda$ coefficients are nonnegative hyperparameters. For dense softmax routing, $\lambda_{\mathrm{load}}$ can be set to zero because every expert has nonzero load by construction.

\section{Experiments}

We evaluate \modelname{} on node classification across eleven benchmark graphs spanning a wide range of scales, homophily regimes, and feature dimensionalities. Our experiments are designed to answer four questions: (i) does joint node--channel adaptive filtering via a small routed expert bank improve over strong \PPGs? (ii) can \modelname{} match the performance of channel-/node-adaptive \MPGs, and graph-MoE baselines? (iii) does the advantage carry over to large graphs where \PPGs are practically necessary? and (iv) which parts of the design are most responsible for the gain, including node adaptivity, channel adaptivity, routing design, diffusion operator choice, and auxiliary regularization?

\subsection{Experimental setup}
\label{sec:exp-setup}

\paragraph{Datasets.}
We use eight small- to medium-scale graphs covering both homophilic and heterophilic regimes together with three large-scale graphs on which the \PPG formulation is most relevant. The details of the dataset and training protocols are provided in Appendix~\ref{app:dataset}.

\paragraph{Baselines.}
We compare against four groups: (i) classical \MPGs; (ii) adaptive spectral \MPGs, (iii) graph-MoE GNNs and (iv) pre-propagation GNNs. The baseline details are provided in Appendix~\ref{app:baseline}. Matched-Chebyshev operator variants for the \PPG baselines are reported separately in Appendix~\ref{app:cheb_kernel_ablation}. Detailed hyperparameter settings are reported in Appendix~\ref{app:leaderboard_hparams}.

\subsection{Main results}
\label{sec:main-results}

\begin{table}[ht]
  \centering
  \captionsetup{width=\textwidth}
  \caption{Averaged node classification accuracy (\%) $\pm$ std over $10$ runs on four homophilic graphs. We highlight the top \textcolor{Green}{\textbf{first}}, \textcolor{NavyBlue}{\textbf{second}}, and \textcolor{Orange}{\textbf{third}} results per dataset.}
  
  \label{tab:main_homo}
  \scalebox{0.88}{
  \begin{tabular}[t]{lcccc}
  \toprule
     & coauthor-cs & coauthor-physics & amazon-photo & amazon-computer \\
  \midrule
  GCN         & $92.92 \pm 0.12$ & $96.18 \pm 0.07$ & $92.70 \pm 0.20$ & $89.65 \pm 0.52$ \\
  GraphSAGE   & $93.91 \pm 0.13$ & $96.49 \pm 0.06$ & $94.59 \pm 0.14$ & $\textcolor{Orange}{\mathbf{91.20}} \pm 0.29$ \\
  GAT         & $93.61 \pm 0.14$ & $96.17 \pm 0.08$ & $93.87 \pm 0.11$ & $90.78 \pm 0.13$ \\
  GPRGNN      & $95.13 \pm 0.09$ & $96.85 \pm 0.08$ & $94.49 \pm 0.14$ & $89.32 \pm 0.29$ \\
  ChebNetII   & $\textcolor{NavyBlue}{\mathbf{95.87}} \pm 0.07$ & $\textcolor{NavyBlue}{\mathbf{97.21}} \pm 0.03$ & $90.70 \pm 0.13$ & $90.22 \pm 0.16$ \\
  \midrule
  ACMGCN      & $95.67 \pm 0.16$ & $96.96 \pm 0.07$ & $94.97 \pm 0.29$ & $90.26 \pm 0.19$ \\
  JacobiConv  & $95.06 \pm 0.08$ & $97.09 \pm 0.02$ & $94.70 \pm 0.15$ & $90.47 \pm 0.15$ \\
  ADC         & $95.45 \pm 0.09$ & $97.03 \pm 0.06$            & $94.64 \pm 0.13$ & $90.07 \pm 0.12$ \\
  DSF         & $94.78 \pm 0.10$ & $96.72 \pm 0.33$ & $94.43 \pm 0.14$ & $84.25 \pm 15.24$ \\
  NFGNN       & $94.72 \pm 0.21$ & $96.94 \pm 0.14$ & $80.80 \pm 29.28$ & $58.51 \pm 27.00$ \\
  AGDN-HA     & $\textcolor{Orange}{\mathbf{95.81}} \pm 0.11$ & $\textcolor{Orange}{\mathbf{97.11}} \pm 0.05$ & $\textcolor{NavyBlue}{\mathbf{96.75}} \pm 0.15$ & $\textcolor{NavyBlue}{\mathbf{91.63}} \pm 0.43$ \\
  AGDN-HC     & $\textcolor{Green}{\mathbf{96.04}} \pm 0.10$ & $97.05 \pm 0.07$ & $\textcolor{Green}{\mathbf{96.78}} \pm 0.11$ & $\textcolor{Green}{\mathbf{91.70}} \pm 0.20$ \\
  \midrule
  GMoE        & $94.45 \pm 0.19$ & $96.52 \pm 0.07$ & $95.91 \pm 0.18$ & $88.57 \pm 0.44$ \\
  Mowst       & $95.35 \pm 0.10$ & $96.60 \pm 0.05$ & $\textcolor{Orange}{\mathbf{96.36}} \pm 0.16$ & $90.87 \pm 0.11$ \\
  NodeMoE     & $95.38 \pm 0.26$ & $\textcolor{Green}{\mathbf{97.34}} \pm 0.04$ & $95.19 \pm 0.55$ & $89.66 \pm 1.47$ \\
  \midrule
  SIGN        & $93.89 \pm 0.40$ & $95.47 \pm 0.22$ & $90.46 \pm 0.69$ & $84.17 \pm 0.34$ \\
  HOGA        & $94.62 \pm 0.25$ & $95.93 \pm 0.28$ & $91.88 \pm 0.71$ & $84.54 \pm 0.76$ \\
  GAMLP       & $94.78 \pm 0.27$ & $96.25 \pm 0.20$ & $91.22 \pm 0.89$ & $86.76 \pm 0.32$ \\
  \midrule
  \modelname (Ours) & $95.61 \pm 0.27$ & $97.10 \pm 0.08$ & $95.88 \pm 0.22$ & $90.68 \pm 0.38$ \\
  \bottomrule
  \end{tabular}
  }
  \vspace{-8pt}
\end{table}

\begin{table}[ht]
  \centering
  \captionsetup{width=\textwidth}
  \caption{Averaged node classification results over $10$ runs on four heterophilic graphs. Accuracy (\%) is reported for amazon-ratings and roman-empire; ROC-AUC is reported for tolokers and minesweeper. We highlight the top \textcolor{Green}{\textbf{first}}, \textcolor{NavyBlue}{\textbf{second}}, and \textcolor{Orange}{\textbf{third}} results per dataset.}
  \label{tab:main_hetero}
  \scalebox{0.88}{
  \begin{tabular}[t]{lcccc}
  \toprule
     & amazon-ratings & tolokers & minesweeper & roman-empire \\
  \midrule
  GCN         & $48.70 \pm 0.63$ & $83.64 \pm 0.67$ & $89.75 \pm 0.52$ & $73.69 \pm 0.74$ \\
  GraphSAGE   & $53.63 \pm 0.39$ & $82.43 \pm 0.44$ & $\textcolor{NavyBlue}{\mathbf{93.51}} \pm 0.57$ & $\textcolor{NavyBlue}{\mathbf{85.74}} \pm 0.67$ \\
  GAT         & $52.70 \pm 0.62$ & $83.78 \pm 0.43$ & $\textcolor{Green}{\mathbf{93.91}} \pm 0.35$ & $\textcolor{Green}{\mathbf{88.75}} \pm 0.41$ \\
  GPRGNN      & $44.88 \pm 0.34$ & $72.94 \pm 0.97$ & $86.24 \pm 0.61$ & $64.85 \pm 0.27$ \\
  ChebNetII   & $51.63 \pm 0.39$ & $78.49 \pm 0.23$ & $81.39 \pm 0.25$ & $74.67 \pm 0.16$ \\
  \midrule
  ACMGCN      & $53.06 \pm 0.36$ & $82.42 \pm 0.59$ & $90.79 \pm 0.55$ & $73.15 \pm 0.60$ \\
  JacobiConv  & $43.77 \pm 0.15$ & $78.28 \pm 0.13$ & $82.26 \pm 2.75$ & $74.57 \pm 0.21$ \\
  ADC         & $49.31 \pm 0.37$ & $80.83 \pm 0.68$ & $83.75 \pm 3.23$ & $78.60 \pm 0.15$ \\
  DSF         & $49.92 \pm 0.41$ & $79.13 \pm 0.51$ & $84.49 \pm 0.35$ & $72.83 \pm 0.71$ \\
  NFGNN       & $47.99 \pm 0.50$ & $79.08 \pm 0.33$ & $85.83 \pm 0.49$ & $75.36 \pm 0.53$ \\
  AGDN-HA     & $53.93 \pm 0.21$ & $81.33 \pm 0.38$ & $87.80 \pm 0.79$ & $82.41 \pm 0.35$ \\
  AGDN-HC     & $\textcolor{Green}{\mathbf{54.32}} \pm 0.43$ & $82.64 \pm 0.38$ & $87.58 \pm 0.52$ & $\textcolor{Orange}{\mathbf{84.55}} \pm 0.58$ \\
  \midrule
  GMoE        & $52.24 \pm 0.31$ & $80.92 \pm 1.03$ & $87.33 \pm 2.46$ & $81.96 \pm 0.35$ \\
  Mowst       & $\textcolor{NavyBlue}{\mathbf{54.29}} \pm 0.31$ & $80.86 \pm 0.25$ & $87.39 \pm 0.47$ & $81.82 \pm 0.57$ \\
  NodeMoE     & $51.33 \pm 0.95$ & $79.05 \pm 0.08$ & $81.33 \pm 2.15$ & $73.99 \pm 2.49$ \\
  \midrule
  SIGN        & $\textcolor{Orange}{\mathbf{54.07}} \pm 0.72$ & $\textcolor{Orange}{\mathbf{84.13}} \pm 0.99$ & $90.71 \pm 0.56$ & $80.01 \pm 0.50$ \\
  HOGA        & $51.56 \pm 0.26$ & $78.10 \pm 0.75$ & $90.53 \pm 0.66$ & $79.39 \pm 0.56$ \\
  GAMLP       & $52.20 \pm 0.40$ & $\textcolor{Green}{\mathbf{85.07}} \pm 0.76$ & $90.47 \pm 0.66$ & $78.87 \pm 0.65$ \\
  \midrule
  \modelname (Ours) & $53.55 \pm 0.81$ & $\textcolor{NavyBlue}{\mathbf{85.06}} \pm 0.76$ & $\textcolor{Orange}{\mathbf{92.29}} \pm 0.49$ & $82.79 \pm 0.55$ \\
  \bottomrule
  \end{tabular}
  }
  \vspace{-8pt}
\end{table}

\begin{table}[ht]
  \centering
  \caption{Averaged node classification accuracy (\%) $\pm$ std on large-scale graphs. For ChebNetII pre-propagation is performed once (no per-epoch message passing), following their original formulations. We highlight the top \textcolor{Green}{\textbf{first}}, \textcolor{NavyBlue}{\textbf{second}}, and \textcolor{Orange}{\textbf{third}} per dataset.}
  \label{tab:main_large}
  \scalebox{0.9}{
  \begin{tabular}[t]{lccc}
  \toprule
     & pokec & ogbn-products & ogbn-papers100M \\
  \midrule
  GCN         & $75.45 \pm 0.17$ & $75.64 \pm 0.21$ &$ 63.29 \pm 0.19$ \\
  GraphSAGE         & $78.40 \pm 0.45$ & $78.40 \pm 0.45$ & $65.79 \pm 0.14$  \\
  GPRGNN      & $78.83 \pm 0.05$ & $79.76 \pm 0.59$ & $65.89 \pm 0.35$ \\
  LINKX       & $82.04 \pm 0.07$ & $71.59 \pm 0.71$ & $56.23 \pm 0.27$ \\
  \midrule
  ChebNetII   & $\textcolor{NavyBlue}{\mathbf{82.33}} \pm 0.28$ & $75.10 \pm 0.24$ & $\textcolor{NavyBlue}{\mathbf{67.18}} \pm 0.32$ \\
  SIGN        & $81.02 \pm 0.33$ & $\textcolor{Orange}{\mathbf{80.52}} \pm 0.16$ & $66.06 \pm 0.19$ \\
  HOGA        & $\textcolor{Orange}{\mathbf{82.11}} \pm 0.25$ & $80.36 \pm 0.21$ & $\textcolor{Orange}{\mathbf{66.86}} \pm 0.11$ \\
  GAMLP       & $78.25 \pm 0.60$ & $\textcolor{NavyBlue}{\mathbf{81.43}} \pm 0.18$ & $66.45 \pm 0.14$ \\
  \midrule
  \modelname (Ours) & $\textcolor{Green}{\mathbf{83.87}} \pm 0.04$ & $\textcolor{Green}{\mathbf{82.36}} \pm 0.25$ & $\textcolor{Green}{\mathbf{67.57}} \pm 0.15$ \\
  \bottomrule
  \end{tabular}
  }
  \vspace{-8pt}
\end{table}

\paragraph{Overall comparison with \PPG baselines.}
Tables~\ref{tab:main_homo}--\ref{tab:main_large} show that \modelname{} is a robust alternative to selecting a different hop aggregator for each dataset. Among the three \PPG baselines, SIGN, HOGA, and GAMLP each win on different benchmarks, so no single hand-designed aggregator is uniformly best. In contrast, \modelname{} improves the average test score over the strongest of these baselines by $+1.53$ points and wins on 9 of 11 datasets. On the remaining two, it is only $0.01$ ROC-AUC points behind GAMLP on \texttt{tolokers} and trails SIGN by $0.52$ accuracy points on \texttt{amazon-ratings}. This supports learned node--channel filter routing as a robust default mechanism, rather than manually tuning among MLP aggregation, hop attention, and adaptive hop fusion.


\paragraph{Homophilic graphs.}
Table~\ref{tab:main_homo} reports results on the four homophilic benchmarks. These datasets are highly saturated at the top: among the five strongest methods in each column, the gap between first and fifth is at most $0.92$ points. In this regime, the best overall results are achieved by models with node- or channel-adaptive propagation, especially the two AGDN variants, which take three of the four first-place entries. \modelname{} remains competitive with these stronger MP-GNN and graph-MoE baselines: it ranks fourth or fifth on 3 out of 4 benchmarks and 6th on the remaining one, trailing the best by at most $1.02$ points.

The main advantage of \modelname{} is clearer within the \PPG block. Compared with the strongest among SIGN, HOGA, and GAMLP on each dataset, \modelname{} improves by $+2.4$ points on average. 
Thus, joint node--channel routing gives a consistent gain over manually chosen \PPG hop aggregators on homophilic datasets. Within the graph-MoE block, \modelname{} also compares favorably: it beats GMoE on three of four datasets and is within $0.03$ points on the remaining one, beats NodeMoE on three of four, and splits wins with Mowst. Overall, the homophilic results suggest that \modelname{} does not uniformly dominate specialized MP-GNNs in small-graph settings, but it substantially strengthens the PP-GNN regime while remaining competitive with adaptive message-passing and graph-MoE alternatives.

\paragraph{Heterophilic graphs.}
Table~\ref{tab:main_hetero} shows that heterophilic benchmarks separate the baselines more clearly than the homophilic ones. \modelname{} is second on \texttt{tolokers} ($85.06$ ROC-AUC), only $0.01$ behind GAMLP, third on \texttt{minesweeper} ($92.29$ ROC-AUC), and fourth on \texttt{roman-empire} ($82.79$). On \texttt{roman-empire} and \texttt{minesweeper}, spatial \MPGs such as GAT and GraphSAGE remain particularly strong, indicating that \modelname{} does not uniformly dominate all message-passing baselines in small heterophilic settings. Its advantage is clearer within the \PPG regime: compared with the strongest among SIGN, HOGA, and GAMLP on each dataset, \modelname{} wins by $+1.58$ on \texttt{minesweeper} and $+2.78$ on \texttt{roman-empire}, is virtually tied on \texttt{tolokers}, and trails only on \texttt{amazon-ratings}. Thus, among \PPG-style models, joint node--channel routing avoids the need to manually choose between MLP-based, hop-attention, and adaptive-fusion aggregators for different heterophilic graphs. 
Compared with graph-MoE baselines, \modelname{} is also favorable overall. Mowst is stronger on \texttt{amazon-ratings}, but \modelname{} improves over the best graph-MoE baseline by $+4.14$ on \texttt{tolokers}, $+4.90$ on \texttt{minesweeper}, and $+0.83$ on \texttt{roman-empire}; NodeMoE trails \modelname{} on all four datasets by $2.22$ to $10.96$ points. Together with the homophilic results, these comparisons suggest that the benefit of \modelname{} is not merely a larger hop-feature aggregator, but a structured way to learn node- and channel-adaptive graph filters while preserving the pre-propagation training path.

\paragraph{Large graphs.}
Table~\ref{tab:main_large} shows the same trend at large scale: \modelname{} is the \emph{best} method on all three large graphs --- $82.36$ on \texttt{ogbn-products} ($+0.93$ over the strongest \PPG baseline GAMLP without label propagation), $67.57$ on \texttt{ogbn-papers100M} ($+0.71$ over HOGA under the same no-label-reuse protocol), and $83.87$ on \texttt{pokec} ($+1.76$ over HOGA). These gains are obtained while keeping all graph-dependent computation in preprocessing.

\subsection{Ablations}
\label{sec:ablation}

Three families of ablations are reported in full in the appendix. \textbf{Router type} (Appendix~\ref{app:gate_ablation}): node-only beats channel-only on all eight small datasets, and joint routers match or outperform both single-axis variants on seven of eight. \textbf{Operator} (Appendix~\ref{app:cheb_kernel_ablation}): a matched Chebyshev basis helps \PPG baselines on homophilic graphs but hurts them on heterophilic graphs. \textbf{Loss terms} (Appendix~\ref{app:loss_ablation}): smoothness loss is the most accuracy-sensitive regularizer with diversity loss having a smaller but consistent effect.

\section{Conclusion}
\label{sec:conclusion}

\modelname{} routes a compact, shared bank of learnable Chebyshev filter experts jointly over nodes and channels while preserving the preprocessing-only training path of \PPGs. Across 11 benchmarks, it improves over the strongest SIGN/HOGA/GAMLP baseline by $+1.53$ test-score points on average, wins on nine datasets, and ranks first on all three web-scale graphs. Together with the ablations, these results show that both axes of adaptivity are important: node-adaptive routing captures graph-position-dependent filtering needs, while channel-adaptive routing allows different feature dimensions to select different spectral responses. This supports joint node--channel filter routing as a robust alternative to manually choosing or tuning hop aggregators for each dataset.

\bibliography{reference_NeurIPS}
\bibliographystyle{plainnat}

\appendix
\section{Extended Background and Related Work}
\label{app:extended_bg_related_work}

This section expands the compact background in Sec.~\ref{sec:bg_related_work}. It is intended for readers less familiar with pre-propagation GNNs (\PPGs{}). The key distinction is where graph-dependent computation occurs: message-passing GNNs (\MPGs) repeatedly aggregate over the graph during training, whereas \PPGs{} amortize graph propagation into a preprocessing stage and train a dense predictor over cached graph-diffused features.

\subsection{Message passing and the scaling bottleneck}
\label{app:bg_mpgnn_scaling}

A generic message-passing layer updates node $i$ by aggregating messages from its neighbors:
\begin{equation}
\mathbf{h}_i^{(\ell+1)}
=
\phi^{(\ell)}\!\left(
\mathbf{h}_i^{(\ell)},
\mathrm{AGG}\Big(\{\psi^{(\ell)}(\mathbf{h}_i^{(\ell)},\mathbf{h}_j^{(\ell)},\mathbf{e}_{ij}) : j\in\mathcal{N}(i)\}\Big)
\right),
\label{eq:app_bg_mp_layer}
\end{equation}
where $\mathrm{AGG}$ is permutation invariant and $\phi^{(\ell)}$ and $\psi^{(\ell)}$ are learnable functions~\citep{gilmer2017neural}. GCN, GraphSAGE, and GAT instantiate this template with normalized aggregation, neighborhood sampling, and attention-based aggregation, respectively~\citep{kipf2016gcn,hamilton2017sage,velickovic2017gat}. The strength of message passing is that propagation and transformation are interleaved: later aggregations operate on task-adapted hidden states. The cost is that training remains graph dependent. After $L$ layers, a target node may depend on an $L$-hop neighborhood, so mini-batch training must sample, gather, or materialize this neighborhood repeatedly.

\subsection{The PP-GNN computational contract}
\label{app:bg_ppgnn_contract}

\PPGs{} decouple graph propagation from learnable transformation. Let $\mathbf{X}\in\mathbb{R}^{N\times F}$ be node features and let $\mathbf{S}_1,\ldots,\mathbf{S}_J\in\mathbb{R}^{N\times N}$ be fixed graph diffusion operators derived from topology. For hop budget $K$, preprocessing constructs
\begin{equation}
\mathbf{Z}
=
\big[
\mathbf{Z}_{1,0}\,\Vert\,\mathbf{Z}_{1,1}\,\Vert\,\cdots\,\Vert\,\mathbf{Z}_{J,K}
\big],
\qquad
\mathbf{Z}_{j,k}=\mathbf{S}_j^k\mathbf{X},
\qquad
\mathbf{Z}\in\mathbb{R}^{N\times F_{\mathrm{eff}}},
\label{eq:app_bg_ppg_preproc}
\end{equation}
where $\mathbf{S}_j^0\mathbf{X}=\mathbf{X}$ and $F_{\mathrm{eff}}=J(K+1)F$. The training stage applies a dense model row-wise,
\begin{equation}
\mathbf{o}_i=f_\theta(\mathbf{z}_i),
\qquad
\hat{\mathbf{y}}_i=\mathrm{softmax}(\mathbf{o}_i).
\label{eq:app_bg_ppg_train}
\end{equation}
Because $\mathbf{z}_i$ already contains the propagated information available to node $i$, a mini-batch consists of rows of a static tensor rather than a sampled subgraph. Thus, after preprocessing, \PPG{} training uses dense matrix operations and can naturally mini-batch over nodes. This is the strict sense in which we use the term \PPG{} in this paper: graph-dependent propagation is not performed inside the training loop.

Common choices of $\mathbf{S}_j$ include normalized adjacency, random-walk transition matrices, and truncated or approximate personalized-PageRank-style diffusions~\citep{wu2019sgc,frasca2020sign,gasteiger2018ppnp,bojchevski2020pprgo,zhang2022gamlp}. Preprocessing costs sparse matrix--dense matrix multiplications, but this cost can be amortized over many epochs, random seeds, and hyperparameter trials when the graph, operator, and input features are fixed. The main systems tradeoff is input expansion: increasing the hop budget, number of operators, or feature dimension enlarges $F_{\mathrm{eff}}$ and therefore the cached tensor~\citep{yue2025graph}.

\subsection{Representative PP-GNN models}
\label{app:bg_representative_ppgnns}

Several existing models instantiate the template in Eqs.~\eqref{eq:app_bg_ppg_preproc}--\eqref{eq:app_bg_ppg_train}. SGC removes intermediate nonlinearities from GCN and trains a linear classifier on a single propagated feature matrix,
\begin{equation}
\hat{\mathbf{Y}}_{\mathrm{SGC}}
=
\mathrm{softmax}\big(\widehat{\mathbf{A}}^{K}\mathbf{X}\mathbf{W}\big),
\label{eq:app_bg_sgc}
\end{equation}
showing that on many homophilic benchmarks much of the benefit of GCN can come from propagation itself~\citep{wu2019sgc}. SIGN precomputes multiple diffused feature matrices and applies dense transformations to their concatenation,
\begin{equation}
\mathbf{H}_{\mathrm{SIGN}}
=
\sigma\!\left(
\mathbf{X}\mathbf{W}_0
\,\Vert\,
(\mathbf{S}_1\mathbf{X})\mathbf{W}_1
\,\Vert\,
\cdots
\,\Vert\,
(\mathbf{S}_J\mathbf{X})\mathbf{W}_J
\right),
\qquad
\hat{\mathbf{Y}}=\mathrm{MLP}(\mathbf{H}_{\mathrm{SIGN}}).
\label{eq:app_bg_sign}
\end{equation}
In the filter view, SIGN learns channel-dependent combinations of cached diffusion features, but the same learned transformations are applied row-wise to every node~\citep{frasca2020sign}. GAMLP learns adaptive fusion over multi-hop feature and label propagations~\citep{zhang2022gamlp}; SAGN uses hop attention together with self-label-enhanced training~\citep{sun2025sagnsle}; and HOGA treats hop features as a sequence and applies gated multi-head attention over the hop dimension~\citep{deng2024hoga}. These methods are more expressive than a plain concatenate-and-MLP head, but their gains are not uniform across datasets, motivating an analysis of the type of adaptivity each aggregator induces.

Related decoupled architectures, such as APPNP, PPRGo, and GPR-GNN, also separate parts of propagation and transformation and have a clear graph-filter interpretation~\citep{gasteiger2018ppnp,bojchevski2020pprgo,chien2020gprgnn}. They do not always satisfy the strict preprocessing-only contract used here, because propagation may be applied to predictions or hidden states after learnable transformations. Nevertheless, they provide important context for the broader decoupled-GNN and graph-filter literature.

\subsection{Diffusion bases as graph filters}
\label{app:bg_filter_view}

For an undirected graph, let $\mathbf{L}=\mathbf{I}-\mathbf{D}^{-1/2}\mathbf{A}\mathbf{D}^{-1/2}=\mathbf{U}\boldsymbol{\Lambda}\mathbf{U}^{\top}$ be the normalized Laplacian. A spectral graph filter with response $g$ acts on a graph signal $\mathbf{x}$ as
\begin{equation}
g(\mathbf{L})\mathbf{x}
=
\mathbf{U}g(\boldsymbol{\Lambda})\mathbf{U}^{\top}\mathbf{x}.
\label{eq:app_bg_spectral_filter}
\end{equation}
Polynomial filters avoid explicit eigendecomposition by writing
\begin{equation}
g(\mathbf{L})\approx\sum_{k=0}^{K}\alpha_k p_k(\mathbf{L}),
\label{eq:app_bg_poly_filter}
\end{equation}
where $p_k$ may be monomials, Chebyshev polynomials, Jacobi polynomials, or another basis~\citep{defferrard2016convolutional,he2022chebnetii,wang2022powerful}. This is the bridge between \PPGs{} and the graph-filter perspective: the cached hop features span a set of candidate graph-filter responses, and the dense head determines how those basis elements are combined.

In \modelname{}, the preprocessing basis is Chebyshev. With a rescaled Laplacian $\widetilde{\mathbf{L}}$ whose spectrum lies in $[-1,1]$, the Chebyshev basis is generated by
\begin{equation}
\mathbf{B}_{0}=\mathbf{X},
\qquad
\mathbf{B}_{1}=\widetilde{\mathbf{L}}\mathbf{X},
\qquad
\mathbf{B}_{k}=2\widetilde{\mathbf{L}}\mathbf{B}_{k-1}-\mathbf{B}_{k-2}.
\label{eq:app_bg_cheb_recurrence}
\end{equation}
Each expert has coefficients $\boldsymbol{\alpha}_m$ over this basis, and routing produces an effective node--channel filter
\begin{equation}
\boldsymbol{\beta}_{i,f}
=
\sum_{m=1}^{M}G_{i,m,f}\boldsymbol{\alpha}_m.
\label{eq:app_bg_filtermoe_beta}
\end{equation}
This gives each node--channel pair an adaptive filter while constraining all effective filters to be mixtures of a compact shared expert bank.

\subsection{Adaptive filtering and graph \MoE{} models}
\label{app:bg_adaptive_moe}

Adaptive graph-filtering work relaxes the assumption that one global propagation rule should be shared everywhere. Channel-adaptive methods learn multiple diffusion scales, per-channel diffusion radii, adaptive hop weights, polynomial coefficients, or mixtures of propagation operators~\citep{atwood2016diffusion,zhao2021adaptive,dong2021adagnn,wang2022powerful,eliasof2023improving}. Node-adaptive methods make aggregation or filtering depend on the node or local context, including attention-based spatial GNNs and node-oriented spectral filters~\citep{velickovic2017gat,luan2022revisiting,sun2020adaptive,zheng2023node,guo2023graph,gama2021nodevariant}. These models motivate adaptivity, but most either operate inside message passing or adapt only one axis at a time.

Mixture-of-experts models offer a complementary mechanism for conditional specialization~\citep{jacobs1991adaptive,shazeer2017outrageously,fedus2022switch}. Recent graph \MoE{} models route among graph encoders, aggregation modules, or node-wise experts~\citep{park2024graphdive,wang2023gmoe,park2023mowst,chen2024nodemoe,wang2024gnnmoe}. \modelname{} differs in where routing is applied: its experts are not full GNN backbones, but learnable Chebyshev graph filters over a cached preprocessing basis. The router therefore selects spectral responses directly at the node--channel level, aligning the \MoE{} mechanism with the graph-filter interpretation of \PPGs{}.

\section{Dataset details}
\label{app:dataset}
\begin{table*}[t]
  \centering
  \small
  \captionsetup{width=\textwidth}
  \caption{Dataset statistics. $N$ and $E$ denote the number of nodes and edges (undirected edges counted once; directed graphs count directed edges).
  Split: ``fixed'' denotes the standard public split; ``random'' denotes averages over multiple random splits.}
  \vspace{-6pt}
  \label{tab:dataset_stats}
  \setlength{\tabcolsep}{3pt}
  \renewcommand{\arraystretch}{0.95}

  \begin{tabular}{lrrrrrrc}
    \toprule
    Dataset & $N$ & $E$ & Feat.\ dim & \#Classes & Train/Val/Test & Split & Type \\
    \midrule
    \texttt{roman-empire}   & $22,662$    & $32,927$     & $300$   & $18$ & 50/25/25 & fixed & Hetero \\
    \texttt{amazon-ratings} & $24,492$    & $93,050$     & $300$   & $5$  & 50/25/25 & fixed & Hetero \\
    \texttt{minesweeper}    & $10,000$    & $39,402$     & $7$     & $2$  & 50/25/25 & fixed & Hetero \\
    \texttt{tolokers}       & $11,758$    & $519,000$    & $10$    & $2$  & 50/25/25 & fixed & Hetero \\
    \texttt{pokec}          & $1,632,803$ & $30,622,564$ & $65$    & $2$  & 50/25/25 & fixed & Hetero \\
    \midrule
    \texttt{amazon-photo}    & $7,650$   & $119,081$   & $745$   & $8$  & 60/20/20 & random & Homo \\
    \texttt{amazon-computer} & $13,752$  & $245,861$   & $767$   & $10$ & 60/20/20 & random & Homo \\
    \texttt{coauthor-cs}     & $18,333$  & $81,894$    & $6,805$ & $15$ & 60/20/20 & random & Homo \\
    \texttt{coauthor-physics}& $34,493$  & $247,962$   & $8,415$ & $5$  & 60/20/20 & random & Homo \\
    \texttt{ogbn-products}      & $2,449,029$ & $61,859,140$ & $100$   & $47$ & 0.08/0.02/0.9 & fixed & Homo \\
    \texttt{ogbn-papers100M} & $111,059,956$ & $1,615,685,872$ & $128$ & $172$ & 0.78/0.08/0.14 & fixed & Homo \\ 
    
    \bottomrule
  \end{tabular}
  \vspace{-10pt}
\end{table*}

The $11$ datasets used in our experiments are summarized in Table~\ref{tab:dataset_stats}.

\textbf{Split protocol.} We follow the standard split protocol for each dataset.
For the four homophily datasets \texttt{\seqsplit{amazon-computer}}, \texttt{\seqsplit{amazon-photo}}, \texttt{\seqsplit{coauthor-cs}}, and \texttt{\seqsplit{coauthor-physics}}, we report mean $\pm$ standard deviation over random splits.
All other datasets use their fixed public splits.

\section{Hardware settings}
\label{app:hardware}
For the training efficiency study, we use a Linux server with a single NVIDIA A100 GPU (CUDA 12.6).
We use the scalable \PPG training framework of \citet{yue2025graph}.

\section{Baseline details}
\label{app:baseline}
We compare against four groups of baselines. First are classical \MPGs: GCN, GraphSAGE, GAT, GPRGNN, and ChebNetII \citep{kipf2016gcn,hamilton2017sage,velickovic2017gat,chien2020gprgnn,he2022chebnetii}. Second are node- or channel-adaptive spectral and spatial \MPG baselines: ACM-GCN, JacobiConv, ADC, DSF, NFGNN, AGDN-HA, AGDN-HC \citep{luan2022revisiting,wang2022powerful,zhao2021adaptive,guo2023graph,zheng2023node,sun2020adaptive}. Third are the graph-MoE baselines: GMoE, Mowst, and NodeMoE \citep{wang2023gmoe,park2023mowst, chen2024nodemoe}. Since we cannot find the opensourced codebase for NodeMoE, we reimplement it based on the description from paper. Last are the most relevant \PPG{} backbones: SIGN, HOGA, and GAMLP \citep{frasca2020sign,deng2024hoga,zhang2022gamlp}. In the main tables, these three \PPG{} baselines use their default diffusion operators, namely either random-walk diffusion or symmetric normalized adjacency. We provide additional results for the \PPG baselines with the Chebyshev diffusion basis in Appendix~\ref{app:cheb_kernel_ablation}. The hyperparameter settings for baselines can be referred to Appendix~\ref{app:baseline_hparams}.

\paragraph{Protocol for ChebNetII and GAMLP.}
Two baselines receive a specific protocol to keep the comparison with
FilterMoE clean.  \emph{ChebNetII}~\citep{he2022chebnetii} is trained in two
modes selected by graph scale, matching the protocol of its original paper:
on the three large graphs (\texttt{pokec}, \texttt{ogbn-products},
\texttt{ogbn-papers100M}) the Chebyshev basis is precomputed once and the
model is trained directly on the cached basis without per-epoch message
passing, mirroring the pre-propagation regime of FilterMoE; on the remaining
eight small- and medium-scale graphs it uses the standard per-layer
message-passing recurrence.  \emph{GAMLP}~\citep{zhang2022gamlp} is run
\emph{without} its optional label-reuse component (label propagation plus
reliable-label distillation), since no other \PPG{} backbone or FilterMoE
benefits from a label feedback loop; the reported numbers therefore isolate
GAMLP's hop-attention aggregator and are directly comparable with the other baselines.

\section{Computational Complexity and Empirical Efficiency}
\label{app:complexity}

\modelname{} preserves the main systems advantage of \PPGs{}: all graph-dependent propagation is performed once in preprocessing, and training uses only dense tensor operations. Let $F$ denote the number of input feature channels, $K$ the Chebyshev degree, $M$ the number of filter experts, and $b$ the number of nodes in the current training batch. For full-batch training, set $b=N$.

The one-time preprocessing stage computes the Chebyshev basis
\[
  \mathbf{B}_{k}=T_k(\widetilde{\mathbf L})\mathbf X,
  \qquad k=0,\ldots,K,
\]
using the standard Chebyshev recurrence. This requires $\mathcal{O}(K|\mathcal E|F)$ sparse-matrix--dense-matrix operations and is done once before training. The cached basis requires $\mathcal{O}(N(K+1)F)$ memory.

During training, each expert output is a dense combination of the cached basis tensors:
\[
  \mathbf{H}^{(m)}_{\mathcal B}
  =
  \sum_{k=0}^{K}\alpha_{m,k}\mathbf{B}_{k,\mathcal B},
  \qquad m=1,\ldots,M,
\]
where $\mathcal B$ is the current batch. Computing all expert outputs for the batch costs
\[
  \mathcal{O}\!\left(bM(K+1)F\right).
\]
The node--channel mixture
\[
  \widetilde{\mathbf X}_{i,f}
  =
  \sum_{m=1}^{M}G_{i,m,f}H^{(m)}_{i,f}
\]
costs $\mathcal{O}(bMF)$. The dense prediction head then operates on $\widetilde{\mathbf X}_{\mathcal B}\in\mathbb{R}^{b\times F}$; for a one-hidden-layer head with width $H_{\mathrm{head}}$, its leading cost is $\mathcal{O}(bFH_{\mathrm{head}})$.

Thus, relative to a SIGN-style \PPG{}, the additional per-epoch work is dense: expert recombination, router evaluation, and expert mixing. No per-epoch sparse neighborhood aggregation is reintroduced. The main additional memory during training is the per-batch expert output tensor and gate tensor, each of size $\mathcal{O}(bMF)$. These tensors are batch-local; they need not be materialized for all $N$ nodes at once.

\paragraph{Direct joint MLP router.}
The direct router predicts the full expert--channel logit grid for each node. Let
$\boldsymbol{\phi}_{i}\in\mathbb{R}^{D_{\mathrm{in}}}$ be the node descriptor used by the router. In the default description, $\boldsymbol{\phi}_{i}$ is formed from the concatenated pre-routing expert outputs $\{\mathbf{H}^{(m)}_{i,:}\}_{m=1}^{M}$, optionally augmented with structural encodings. If the raw concatenation is used, then $D_{\mathrm{in}}=MF+D_s$, where $D_s$ is the dimension of the optional structural encoding; if a compressed summary is used, $D_{\mathrm{in}}$ is the summary dimension.

For a depth-$L_g$ MLP with hidden width $D_h$, the direct router computes
\[
  \mathbf L_i
  =
  \operatorname{reshape}
  \!\left(
  \mathrm{MLP}_{\mathrm{joint}}(\boldsymbol{\phi}_i)
  \right)
  \in \mathbb{R}^{M\times F}.
\]
Its leading batch cost is
\[
  \mathcal{O}\!\left(
  bD_{\mathrm{in}}D_h
  +
  b(L_g-2)D_h^2
  +
  bD_hMF
  \right),
\]
for $L_g\ge 2$, plus $\mathcal{O}(bMF)$ for the expert-wise softmax. The corresponding leading parameter count is
\[
  \mathcal{O}\!\left(
  D_{\mathrm{in}}D_h
  +
  (L_g-2)D_h^2
  +
  D_hMF
  \right).
\]
This router is flexible because every expert--channel pair has its own output logit, but the final projection scales with $MF$.

\paragraph{Response-aware two-stage router.}
The response-aware router replaces the direct $M\times F$ output projection with a factorized routing rule. Let $P$ be the number of SLQ spectral grid points and $D$ the expert-key/query dimension. The expert keys are obtained from weighted response sketches,
\[
  \mathbf e_m
  =
  \mathrm{MLP}_{\mathrm{exp}}
  \!\left(
  \mathrm{LN}(\widehat{\mathbf r}_m)
  \right)
  \in\mathbb{R}^{D}.
\]
Computing the sampled responses from the Chebyshev coefficients costs
$\mathcal{O}(MP(K+1))$, and embedding them costs $\mathcal{O}(M(PD+D^2))$ for a one-hidden-layer expert-key MLP. This part is independent of the batch size.

For the node and channel queries, let $D_{\mathrm{node}}$ and $D_{\mathrm{chan}}$ denote the input dimensions of the node-level summary $\boldsymbol{\psi}_i$ and the channel-level summary $\boldsymbol{\psi}_{i,f}$. The query projections cost
\[
  \mathcal{O}\!\left(
  bD_{\mathrm{node}}D
  +
  bFD_{\mathrm{chan}}D
  \right).
\]
The stage-1 dot products
\[
  s^{(1)}_{i,m}
  =
  \langle \mathbf u_i,\mathbf e_m\rangle
\]
cost $\mathcal{O}(bMD)$, and the channel-wise refinement
\[
  \Delta_{i,m,f}
  =
  \langle \mathbf v_{i,f},\mathbf e_m\rangle
\]
costs $\mathcal{O}(bMFD)$. Therefore, the leading cost of the response-aware router is
\[
  \mathcal{O}\!\left(
  MP(K+1)
  +
  M(PD+D^2)
  +
  bD_{\mathrm{node}}D
  +
  bFD_{\mathrm{chan}}D
  +
  bMD
  +
  bMFD
  \right).
\]
If the optional stage-1 top-$K_{\mathrm{route}}$ screening is used, the stage-2 contraction can be reduced from $\mathcal{O}(bMFD)$ to $\mathcal{O}(bK_{\mathrm{route}}FD)$, with $K_{\mathrm{route}}\le M$.

Compared with the direct joint MLP router, the response-aware router trades a learned $D_hMF$ output projection for shared expert keys and dot products. This can reduce parameter count when the key/query dimension $D$ and the summary dimensions are small, but the exact runtime depends on $M$, $F$, $D$, and the chosen summaries. In both router variants, the cost remains dense and batch-local.

\paragraph{Auxiliary losses.}
The auxiliary losses are small compared with the main dense routing and prediction costs. The Chebyshev-coefficient smoothness penalty costs $\mathcal{O}(MK)$. The response-space decorrelation loss costs $\mathcal{O}(MP(K+1)+M^2P)$ if the response sketches are not already available, and only $\mathcal{O}(M^2P)$ when they are reused from the response-aware router. The importance, load, and router $z$-loss terms operate on the gate logits or gate tensor and cost $\mathcal{O}(bMF)$. These terms do not involve graph traversal and therefore do not change the \PPG{} training contract.

\paragraph{Summary of overhead.}
\modelname{} adds a controlled dense overhead to SIGN-style \PPG{} training. It shares the same preprocessing-only graph-computation contract, but during training additionally computes $M$ expert outputs, an expert--channel gate, and a node--channel mixture. This makes it more expensive than SIGN by roughly a factor linear in the number of experts, plus the router cost, while keeping all extra tensors batch-local. Compared with HOGA-style hop attention, \modelname{} avoids quadratic attention over hop pairs and instead uses a compact Chebyshev expert bank with routing costs linear in the number of experts and channels. Thus, the intended tradeoff is clear: FilterMoE spends more dense computation than SIGN to gain joint node--channel adaptivity, but avoids the heavier hop-attention pattern and manual head design of HOGA.

\paragraph{One-time preprocessing overhead.}

Table~\ref{tab:preproc_runtime} separates the offline preprocessing cost of \modelname{} from the online training/evaluation time on the three large graphs. The preprocessing stage produces graph-level cached artifacts used by the method: the Chebyshev basis $\mathbf{B}_{k}=T_k(\widetilde{\mathbf L})\mathbf X$ for $k=0,\ldots,K$, and the SLQ spectral grid $\{(\theta_p,w_p)\}_{p=1}^{P}$ used by the response-aware router and the diversity loss term. These computations touch the full graph once and are then amortized across repeated runs. In contrast, \emph{train/eval wall per run} measures the average wall-clock time of one selected-configuration run after cached features are available, including data loading, training, evaluation, and saving.

For \texttt{pokec} and \texttt{ogbn-products}, preprocessing is a small fraction of the 10-run cost, below $3\%$ in this profile. For \texttt{ogbn-papers100M}, the relative preprocessing share is larger because Chebyshev and SLQ preprocessing must scan the full 111M-node graph, whereas the \PPG training cost is tied to the labeled nodes, which only take about 1.4\% of total nodes. Even in this case, the one-time preprocessing cost is amortized over the 10-seed evaluation and the total wall-clock profile remains dominated by dense training/evaluation rather than repeated sparse propagation.

\begin{table}[ht]
  \centering
  \captionsetup{width=\textwidth}
  \caption{Wall-clock profile of \modelname{} on the three large graphs. Cheb-basis preprocessing materializes $\mathbf{B}_{k}=T_k(\widetilde{\mathbf L})\mathbf X$ for $k=0,\ldots,K$; SLQ constructs the graph-specific spectral grid. \emph{Train/eval wall per run} is averaged over the 10 selected-configuration runs after cached features are available. The 10-run total equals preprocessing plus $10\times$ train/eval wall time.}
  \label{tab:preproc_runtime}
  \scalebox{0.85}{
  \begin{tabular}[t]{lrrrrr}
  \toprule
  Dataset & Nodes & Cheb basis & SLQ (GPU) & Train wall / run & End-to-end ($r{=}10$) \\
  \midrule
  \texttt{pokec}            & 1.6\,M   & 60\,s      & 25\,s      & 1863\,s ($\approx$31\,min)   & $\approx$5.2\,h \\
  \texttt{ogbn-products}    & 2.4\,M   & 103\,s     & 67\,s      & 607\,s ($\approx$10\,min)    & $\approx$1.7\,h \\
  \texttt{ogbn-papers100M}  & 111\,M   & 1185\,s ($\approx$20\,min) & 1276\,s ($\approx$21\,min) & 837\,s ($\approx$14\,min) & $\approx$3.0\,h \\
  \bottomrule
  \end{tabular}
  }
\end{table}

\paragraph{Training/evaluation runtime--accuracy trade-off.}
\begin{figure}[t]
  \centering
\includegraphics[width=\linewidth]{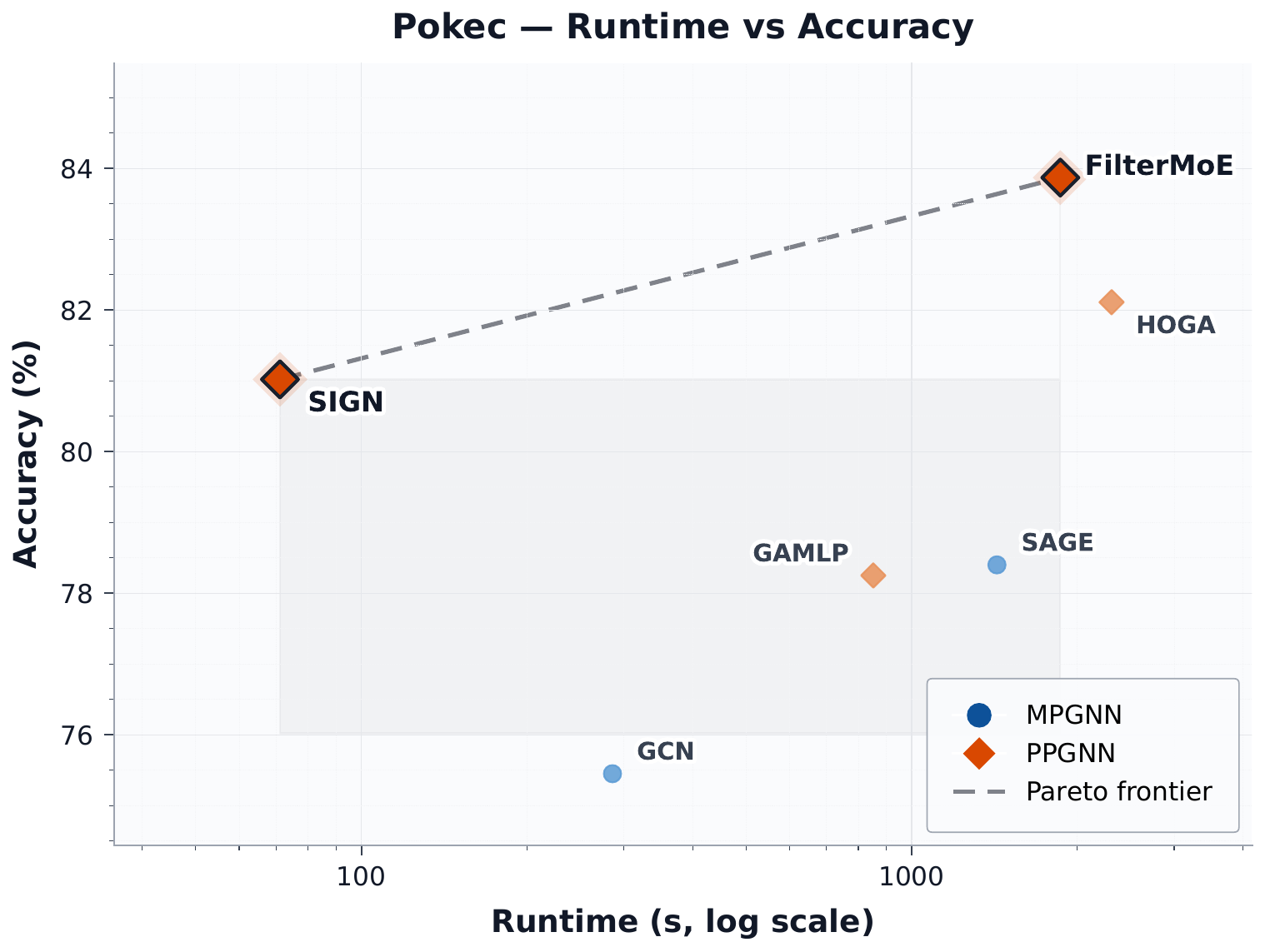}
  \caption{Runtime--accuracy trade-off on \texttt{pokec}. Runtime includes training and evaluation, but excludes one-time preprocessing, and is averaged over 10 runs; Table~\ref{tab:preproc_runtime} reports the preprocessing overhead separately. Markers distinguish message-passing GNNs and PP-GNNs, and the dashed line indicates the Pareto frontier.}
  \label{fig:pokec_runtime_accuracy}
\end{figure}
To complement the asymptotic analysis, Figure~\ref{fig:pokec_runtime_accuracy} reports the runtime--accuracy trade-off on \texttt{pokec} after cached features have been constructed. SIGN remains the fastest \PPG{} baseline in this online training/evaluation regime, but obtains substantially lower accuracy. \modelname{} incurs additional dense computation for expert recombination and node--channel routing, yet lies on the Pareto frontier and achieves the highest accuracy among the plotted methods. HOGA and GAT are slower in this measurement while obtaining lower accuracy than \modelname{}. Together with Table~\ref{tab:preproc_runtime}, this supports the intended trade-off: \modelname{} pays a modest one-time preprocessing cost and additional dense training cost to gain joint node--channel adaptivity, without reintroducing per-epoch sparse message passing.

\section{Router-type ablation}
\label{app:gate_ablation}

To isolate the role of the router, we compare four variants of \modelname{} that use the same Chebyshev expert bank, output head, and training protocol, but differ in how the gate $G_{i,m,f}$ is parameterized. The \emph{Node-only} router produces one expert distribution per node and broadcasts it across feature channels, i.e., $G_{i,m,f}=\pi_{i,m}$. The \emph{Channel-only} router is the symmetric counterpart: it produces one expert distribution per channel and shares it across all nodes, i.e., $G_{i,m,f}=\bar\pi_{m,f}$. Concretely, before training we compute per-channel Chebyshev spectral moments
\[
  \mu_{f,k} \;=\; \frac{\mathbf{x}_f^{\top}T_{k}(\tilde{\mathbf L})\,\mathbf{x}_f}{\|\mathbf{x}_f\|^{2}},
  \qquad k=0,\ldots,K,
\]
which yield a $(F,K{+}1)$ matrix that summarizes how each channel's energy distributes over the graph spectrum (low-pass channels concentrate moments at small $k$, high-pass channels at large $k$). At training time, an MLP plus LayerNorm maps each row $\boldsymbol{\mu}_f$ to a channel embedding $\mathbf{z}_f \in \mathbb{R}^{D}$. Each filter is represented by the same weighted response embedding $\mathbf{e}_m \in \mathbb{R}^{D}$ used in the response-aware router (Eq.~\eqref{eq:filtermoe-expert-key}). The gate logits are then
\[
  L_{m,f} \;=\; \langle \mathbf{e}_{m}, \mathbf{z}_{f}\rangle,
  \qquad
  \bar\pi_{m,f}=\frac{\exp(L_{m,f}/\tau)}{\sum_{m'}\exp(L_{m',f}/\tau)},
\]
i.e., each channel selects the filter whose spectral response best matches its own spectral profile, with no node-level information entering the gate. The \emph{Direct joint MLP} router is the direct router from Sec.~\ref{sec:method}: it maps descriptors of the pre-routing expert outputs $\{\mathbf{H}^{(m)}\}_{m=1}^{M}$, optionally augmented with random-walk structural encodings, to a per-node $M\times F$ expert--channel logit grid. The \emph{Response-aware two-stage} router is the second router from Sec.~\ref{sec:method}: it embeds each expert by its weighted spectral-response sketch, computes a node-level prior over experts, and refines this prior with channel-level residual scores.

\begin{table}[ht]
  \centering
  \captionsetup{width=\textwidth}
  \caption{Router-type ablation for \modelname{} on the four \textbf{homophilic} graphs. Rows compare four parameterizations of the gate $G_{i,m,f}$: \textbf{Node-only} is node-adaptive only and broadcasts one expert distribution across channels; \textbf{Channel-only} is channel-adaptive only and shares one expert distribution across nodes; \textbf{Direct joint MLP} is the direct router in Sec.~\ref{sec:method-routing}, which emits a per-node $M\times F$ expert--channel logit grid; \textbf{Response-aware two-stage} is the router in Sec.~\ref{sec:method-routing} that combines spectral-response expert keys with a node-level prior and channel-level residual refinement. Values are test accuracy mean $\pm$ std over $10$ runs after Optuna tuning. \textbf{Bold} marks the highest mean per column; exact ties in mean are also bold. The heterophilic half is reported in Table~\ref{tab:ablation_gate_hetero}.}
  \label{tab:ablation_gate_homo}
  \scalebox{0.88}{
  \begin{tabular}[t]{lcccc}
  \toprule
   & coauthor-cs & coauthor-physics & amazon-photo & amazon-computer \\
  \midrule
  Node-only                    & $\mathbf{95.61 \pm 0.33}$ & $\mathbf{97.23 \pm 0.05}$ & $95.41 \pm 0.27$ & $90.05 \pm 2.63$ \\
  Channel-only                 & $95.45 \pm 0.45$ & $97.19 \pm 0.13$ & $95.24 \pm 0.28$ & $87.94 \pm 1.66$ \\
  Direct joint MLP             & $\mathbf{95.61 \pm 0.27}$ & $97.10 \pm 0.08$ & $94.81 \pm 0.52$ & $\mathbf{90.68 \pm 0.38}$ \\
  Response-aware two-stage     & $95.44 \pm 0.29$           & $96.14 \pm 0.24$ & $\mathbf{95.88 \pm 0.22}$ & $88.18 \pm 1.95$ \\
  \bottomrule
  \end{tabular}
  }
\end{table}

\begin{table}[ht]
  \centering
  \captionsetup{width=\textwidth}
  \caption{Router-type ablation for \modelname{} on the four \textbf{heterophilic} graphs. The same four gate parameterizations and tuning/evaluation protocol as Table~\ref{tab:ablation_gate_homo} are used. ROC-AUC is reported for \texttt{tolokers} and \texttt{minesweeper}; accuracy is reported otherwise. \textbf{Bold} marks the highest mean per column.}
  \label{tab:ablation_gate_hetero}
  \scalebox{0.88}{
  \begin{tabular}[t]{lcccc}
  \toprule
   & amazon-ratings & tolokers & minesweeper & roman-empire \\
  \midrule
  Node-only                    & $52.13 \pm 1.15$ & $84.20 \pm 0.70$ & $90.84 \pm 0.79$ & $74.67 \pm 2.51$ \\
  Channel-only                 & $51.94 \pm 1.08$ & $84.07 \pm 0.53$ & $89.97 \pm 0.54$ & $73.53 \pm 1.73$ \\
  Direct joint MLP             & $\mathbf{53.55 \pm 0.81}$ & $\mathbf{85.06 \pm 0.76}$ & $\mathbf{92.29 \pm 0.49}$ & $\mathbf{82.62 \pm 0.57}$ \\
  Response-aware two-stage     & $52.74 \pm 1.48$ & $82.23 \pm 1.18$ & $89.86 \pm 0.71$ & $73.74 \pm 0.94$ \\
  \bottomrule
  \end{tabular}
  }
\end{table}

The results show two consistent trends. First, the node-only router matches or exceeds the channel-only router on all eight datasets, with margins from $0.04$ to $2.11$ points. This suggests that, when only one axis of adaptivity is available, node-dependent routing is usually more informative than a node-shared channel gate. Second, a joint router matches or outperforms both single-axis variants on seven of the eight datasets. The only exception is \texttt{coauthor-physics}, where the node-only and channel-only gates already reach $97.23$ and $97.19$, respectively, and the joint routers provide no additional benefit. This is consistent with the broader homophilic results, where several feature-rich graphs are close to saturation.

Among the two joint routers, the direct joint MLP is the most robust in this ablation. It wins or ties on six datasets and is especially strong on heterophilic graphs: compared with the node-only router, it improves by $+1.42$ on \texttt{amazon-ratings}, $+0.86$ on \texttt{tolokers}, $+1.45$ on \texttt{minesweeper}, and $+7.95$ on \texttt{roman-empire}. The response-aware two-stage router gives the best result on \texttt{amazon-photo} and remains competitive on \texttt{amazon-ratings}, but its factorized node-prior-plus-channel-residual form is less reliable on several heterophilic datasets. Taken together, the ablation supports the main architectural conclusion: the important ingredient is joint node--channel routing over a shared learnable Chebyshev expert bank, while the best concrete router parameterization can vary by dataset.

\section{PP-GNN diffusion-operator ablation}
\label{app:cheb_kernel_ablation}

This ablation isolates whether the gains of \modelname{} can be explained simply by changing the preprocessing operator used by existing \PPG{} backbones. For each backbone, we compare two settings. The \emph{original-operator} setting uses the same random-walk-based diffusion operator or normalized-adjacency preprocessing used for the corresponding \PPG{} baseline in the main tables. The \emph{Chebyshev-operator} setting replaces only this preprocessing operator with the Chebyshev basis used by \modelname{}, while keeping the downstream SIGN, HOGA, or GAMLP head unchanged. Importantly, the Chebyshev variants are not evaluated with fixed hyperparameters from the original-operator runs: each variant is independently tuned with a 50-trial Optuna TPE search, and the selected configuration is then evaluated over $10$ random seeds. Thus, the comparison varies only the preprocessing operator while keeping the \PPG{} head design fixed.

\begin{table}[ht]
  \centering
  \captionsetup{width=\textwidth}
  \caption{Diffusion-operator ablation for \PPG{} baselines on the four \textbf{homophilic} graphs. For each backbone, \emph{Orig.} uses the same random-walk-based diffusion operator or normalized-adjacency preprocessing as the corresponding main-table baseline; \emph{Cheb.} replaces only the preprocessing operator with the Chebyshev basis used by \modelname{}. Chebyshev variants are independently tuned with a 50-trial Optuna TPE search and evaluated over $10$ seeds; deltas are relative to the original-operator result.}
  \label{tab:ppgnn_kernel_ablation_homo}
  \resizebox{\textwidth}{!}{
  \begin{tabular}[t]{llcccc}
  \toprule
  Backbone & Operator & coauthor-cs & coauthor-physics & amazon-photo & amazon-computer \\
  \midrule
  SIGN  & Orig. & $93.89 \pm 0.40$ & $95.47 \pm 0.22$ & $90.46 \pm 0.69$ & $84.17 \pm 0.34$ \\
        & Cheb. & $94.86 \pm 0.33$ {\scriptsize $(+0.97)$}
                 & $96.21 \pm 0.14$ {\scriptsize $(+0.74)$}
                 & $93.83 \pm 0.31$ {\scriptsize $(+3.37)$}
                 & $88.91 \pm 0.26$ {\scriptsize $(+4.74)$} \\
  \midrule
  HOGA  & Orig. & $94.62 \pm 0.25$ & $95.93 \pm 0.28$ & $91.88 \pm 0.71$ & $84.54 \pm 0.76$ \\
        & Cheb. & $94.94 \pm 0.43$ {\scriptsize $(+0.32)$}
                 & $96.51 \pm 0.09$ {\scriptsize $(+0.58)$}
                 & $94.40 \pm 0.33$ {\scriptsize $(+2.52)$}
                 & $89.83 \pm 0.39$ {\scriptsize $(+5.29)$} \\
  \midrule
  GAMLP & Orig. & $94.78 \pm 0.27$ & $96.25 \pm 0.20$ & $91.22 \pm 0.89$ & $86.76 \pm 0.32$ \\
        & Cheb. & $94.98 \pm 0.30$ {\scriptsize $(+0.20)$}
                 & $96.60 \pm 0.14$ {\scriptsize $(+0.35)$}
                 & $94.39 \pm 0.43$ {\scriptsize $(+3.17)$}
                 & $89.99 \pm 0.33$ {\scriptsize $(+3.23)$} \\
  \midrule
  \multicolumn{2}{l}{Best Cheb. \PPG{}} & $94.98 \pm 0.30$ {\scriptsize (GAMLP)}
                   & $96.60 \pm 0.14$ {\scriptsize (GAMLP)}
                   & $94.40 \pm 0.33$ {\scriptsize (HOGA)}
                   & $89.99 \pm 0.33$ {\scriptsize (GAMLP)} \\
  \multicolumn{2}{l}{\modelname{} (Cheb.)} & $\mathbf{95.61} \pm 0.27$
                      & $\mathbf{97.10} \pm 0.08$
                      & $\mathbf{95.88} \pm 0.22$
                      & $\mathbf{90.68} \pm 0.38$ \\
  \multicolumn{2}{l}{$\Delta$ vs. best Cheb. \PPG{}} & $\mathbf{+0.63}$ & $\mathbf{+0.50}$ & $\mathbf{+1.48}$ & $\mathbf{+0.69}$ \\
  \bottomrule
  \end{tabular}
  }
\end{table}

\begin{table}[ht]
  \centering
  \captionsetup{width=\textwidth}
  \caption{Diffusion-operator ablation for \PPG{} baselines on the four \textbf{heterophilic} graphs. For each backbone, \emph{Orig.} uses the same random-walk-based diffusion operator or normalized-adjacency preprocessing as the corresponding main-table baseline; \emph{Cheb.} replaces only the preprocessing operator with the Chebyshev basis used by \modelname{}. Chebyshev variants are independently tuned with a 50-trial Optuna TPE search and evaluated over $10$ seeds; deltas are relative to the original-operator result. Accuracy is reported for \texttt{amazon-ratings} and \texttt{roman-empire}; ROC-AUC is reported for \texttt{tolokers} and \texttt{minesweeper}.}
  \label{tab:ppgnn_kernel_ablation_hetero}
  \resizebox{\textwidth}{!}{
  \begin{tabular}[t]{llcccc}
  \toprule
  Backbone & Operator & amazon-ratings & tolokers & minesweeper & roman-empire \\
  \midrule
  SIGN  & Orig. & $54.07 \pm 0.72$ & $84.13 \pm 0.99$ & $90.71 \pm 0.56$ & $80.01 \pm 0.50$ \\
        & Cheb. & $48.72 \pm 0.52$ {\scriptsize $(-5.35)$}
                 & $83.43 \pm 0.94$ {\scriptsize $(-0.70)$}
                 & $91.49 \pm 0.56$ {\scriptsize $(+0.78)$}
                 & $78.12 \pm 0.52$ {\scriptsize $(-1.89)$} \\
  \midrule
  HOGA  & Orig. & $51.56 \pm 0.26$ & $78.10 \pm 0.75$ & $90.53 \pm 0.66$ & $79.39 \pm 0.56$ \\
        & Cheb. & $51.23 \pm 0.52$ {\scriptsize $(-0.33)$}
                 & $83.23 \pm 0.88$ {\scriptsize $(+5.13)$}
                 & $91.86 \pm 0.55$ {\scriptsize $(+1.33)$}
                 & $81.51 \pm 0.51$ {\scriptsize $(+2.12)$} \\
  \midrule
  GAMLP & Orig. & $52.20 \pm 0.40$ & $85.07 \pm 0.76$ & $90.47 \pm 0.66$ & $78.87 \pm 0.65$ \\
        & Cheb. & $48.66 \pm 1.11$ {\scriptsize $(-3.54)$}
                 & $83.82 \pm 0.79$ {\scriptsize $(-1.25)$}
                 & $91.70 \pm 0.60$ {\scriptsize $(+1.23)$}
                 & $72.93 \pm 0.75$ {\scriptsize $(-5.94)$} \\
  \midrule
  \multicolumn{2}{l}{Best Cheb. \PPG{}} & $51.23 \pm 0.52$ {\scriptsize (HOGA)}
                   & $83.82 \pm 0.79$ {\scriptsize (GAMLP)}
                   & $91.86 \pm 0.55$ {\scriptsize (HOGA)}
                   & $81.51 \pm 0.51$ {\scriptsize (HOGA)} \\
  \multicolumn{2}{l}{\modelname{} (Cheb.)} &  $53.55 \pm 0.81$
                      & $\mathbf{85.06} \pm 0.76$
                      & $\mathbf{92.29} \pm 0.49$
                      & $\mathbf{82.79} \pm 0.55$ \\
  \multicolumn{2}{l}{$\Delta$ vs. best Cheb. \PPG{}} & $\mathbf{+2.32}$ & $\mathbf{+1.24}$ & $\mathbf{+0.43}$ & $\mathbf{+1.28}$ \\
  \bottomrule
  \end{tabular}
  }
\end{table}

Tables~\ref{tab:ppgnn_kernel_ablation_homo}--\ref{tab:ppgnn_kernel_ablation_hetero} show that the Chebyshev operator is useful but not sufficient to explain the full result. On homophilic graphs, replacing the original random-walk or normalized-adjacency preprocessing with a tuned Chebyshev basis improves every SIGN/HOGA/GAMLP cell, with an average gain of $+2.12$ test-score points and a maximum gain of $+5.29$ on \texttt{amazon-computer}. However, even after this matched-operator tuning, the best Chebyshev \PPG{} baseline remains below \modelname{} on all four homophilic datasets by $+0.5$ to $+1.48$ points.

The heterophilic results are more mixed. Chebyshev preprocessing improves only five of the twelve backbone--dataset cells and decreases the average score by $0.70$ points, with large drops for SIGN on \texttt{amazon-ratings} and for GAMLP on \texttt{roman-empire}. Nevertheless, \modelname{} still outperforms the best Chebyshev-tuned \PPG{} baseline on every heterophilic dataset, by $+0.43$ to $+2.32$ points. Thus, the advantage of \modelname{} does not come from the Chebyshev preprocessing operator alone. The matched-operator ablation supports the main interpretation: the key gain comes from learning node- and channel-adaptive mixtures over a shared Chebyshev filter bank, rather than from replacing the original \PPG{} diffusion operator with a Chebyshev basis while retaining a fixed SIGN-, HOGA-, or GAMLP-style aggregator.

\section{Loss-term ablation}
\label{app:loss_ablation}

We ablate the auxiliary loss terms in \modelname{} by starting from the selected configuration used in the main results, setting one auxiliary coefficient to zero, and evaluating the modified model over $10$ random seeds without retuning the remaining hyperparameters. The selected configuration should not be interpreted as using every auxiliary loss: during hyperparameter search, each coefficient may be selected as zero. Therefore, when an ablation cell exactly matches the selected-objective result, the corresponding regularization term was already inactive in that configuration, and zeroing it is a no-op. For \texttt{tolokers} and \texttt{minesweeper}, the reported score is ROC-AUC; for all other datasets, it is accuracy. Columns 2 to 6 zero the corresponding filter diversity, importance, load, smoothness, or router $z$-loss coefficient.

\begin{table}[ht]
  \centering
  \captionsetup{width=\textwidth}
  \caption{Loss-term ablation for \modelname{}. Each ablation starts from the selected configuration used in the main results, sets the named auxiliary loss coefficient to zero, and evaluates the modified model over $10$ random seeds. Values are mean test scores (\%) $\pm$ std. The selected configuration may already set some auxiliary coefficients to zero; for such cases we just repeat the value in the main results.}
  \label{tab:loss_ablation}
  \scalebox{0.70}{
  \begin{tabular}[t]{lcccccc}
  \toprule
   & selected objective & no\_diversity & no\_importance & no\_load & no\_smoothness & no\_zloss \\
  \midrule
  coauthor-cs       & $95.61 \pm 0.27$ & $95.60 \pm 0.33$ & $95.61 \pm 0.24$ & $95.61 \pm 0.29$ & $95.55 \pm 0.27$ & $95.50 \pm 0.31$ \\
  coauthor-physics  & $97.10 \pm 0.08$ & $97.09 \pm 0.11$ & $97.10 \pm 0.08$ & $97.10 \pm 0.09$ & $97.07 \pm 0.10$ & $96.99 \pm 0.11$ \\
  amazon-photo      & $95.88 \pm 0.22$ & $95.69 \pm 0.22$ & $95.72 \pm 0.36$ & $95.72 \pm 0.36$ & $95.47 \pm 0.58$ & $95.71 \pm 0.35$ \\
  amazon-computer   & $90.68 \pm 0.38$ & $90.68 \pm 0.38$ & $90.68 \pm 0.38$ & $90.91 \pm 0.17$ & $89.78 \pm 0.48$ & $90.68 \pm 0.38$ \\
  amazon-ratings    & $53.55 \pm 0.81$ & $53.44 \pm 0.80$ & $53.42 \pm 0.84$ & $53.42 \pm 0.84$ & $52.03 \pm 0.66$ & $52.85 \pm 0.52$ \\
  tolokers          & $85.06 \pm 0.76$ & $84.84 \pm 0.56$ & $84.35 \pm 0.57$ & $84.72 \pm 0.79$ & $84.27 \pm 0.69$ & $84.71 \pm 0.49$ \\
  minesweeper       & $92.29 \pm 0.49$ & $91.96 \pm 0.62$ & $92.01 \pm 0.68$ & $92.01 \pm 0.68$ & $91.73 \pm 0.48$ & $92.01 \pm 0.68$ \\
  roman-empire      & $82.79 \pm 0.55$ & $82.28 \pm 0.62$ & $82.66 \pm 0.38$ & $82.66 \pm 0.38$ & $82.26 \pm 0.80$ & $82.66 \pm 0.38$ \\
  \bottomrule
  \end{tabular}
  }
\end{table}

The most consistent regularization effect comes from the Chebyshev-coefficient smoothness penalty. Removing smoothness decreases the mean score on all eight datasets, the largest drop is on \texttt{amazon-ratings}, from $53.55 \pm 0.81$ to $52.03 \pm 0.66$. On the other datasets, the effect is smaller but still consistent, ranging from $0.03$ to $0.90$ points. This suggests that smoothness is most important on noisier or more heterophilic signals, while its effect is modest on saturated homophilic graphs.

The filter diversity loss plays a more auxiliary role. Removing it changes the mean by at most $0.51$ points on eight datasets. 
The MoE balancing losses and router $z$-loss are also dataset-dependent. Exact repeated cells, such as the no\_importance entries on \texttt{coauthor-physics} and the no\_diversity/no\_importance/no\_zloss entries on \texttt{amazon-computer}, indicate that those coefficients were already zero in the selected configuration. When active, these terms have modest effects: on \texttt{amazon-ratings}, removing importance or load lowers the score by $0.13$ points, while removing the $z$-loss lowers it by $0.70$ points; on \texttt{tolokers}, the corresponding drops are $0.71$, $0.34$, and $0.35$ points. Overall, the loss ablation indicates that \modelname{} does not rely on a delicate combination of auxiliary losses. The main architectural gains come from joint node--channel routing over the shared Chebyshev expert bank; smoothness is the most consistent regularizer, while filter diversity loss and MoE balancing losses provide smaller, dataset-dependent effects.

\section{\modelname{} Hyperparameter Configurations}
\label{app:leaderboard_hparams}

For reproducibility, Tables~\ref{tab:hparams_training}--\ref{tab:hparams_loss_rwse} report the selected hyperparameters used for the \modelname{} rows in Tables~\ref{tab:main_homo}, \ref{tab:main_hetero}, and \ref{tab:main_large}. Each configuration is selected by Optuna TPE and then evaluated over $10$ random seeds. 

\paragraph{Shared spectral preprocessing.}
All datasets use the same SLQ setup for the spectral response sketches in Sec.~\ref{sec:method-experts}: $20$ random-vector probes, $50$ Lanczos iterations per probe, and a $P=64$ point weighted spectral grid $\{(\theta_p,w_p)\}_{p=1}^{64}$ on the rescaled graph Laplacian. This grid is used by the response-aware two-stage router when active, and by the filter diversity loss when $\lambda_{\mathrm{div}}>0$.

\begin{table}[ht]
  \centering
  \captionsetup{width=\textwidth}
  \caption{Training hyperparameters for \modelname{}. \emph{drop} and \emph{in.drop} denote dropout and input dropout; \emph{pat.} denotes early-stopping patience. All runs use the same train/validation/test split protocol as the main experiments.}
  \label{tab:hparams_training}
  \scalebox{0.85}{
  \begin{tabular}[t]{lcccccccc}
  \toprule
  Dataset & opt. & lr & wd & drop & in.drop & batch & epochs & pat. \\
  \midrule
  \texttt{coauthor-cs} & adamw & 0.001 & 1.0e-05 & 0.5 & 0.3 & 8000 & 860 & 150 \\
  \texttt{coauthor-physics} & adamw & 0.03 & 0 & 0.5 & 0.3 & 8000 & 200 & 100 \\
  \texttt{amazon-photo} & adamw & 0.02 & 1.0e-05 & 0.4 & 0.1 & 1000 & 110 & 50 \\
  \texttt{amazon-computer} & adamw & 0.008 & 1.0e-04 & 0.5 & 0.1 & 8000 & 200 & 50 \\
  \texttt{amazon-ratings} & adamw & 0.03 & 0 & 0.5 & 0.22 & 1000 & 1000 & 100 \\
  \texttt{tolokers} & adamw & 0.0059 & 0 & 0.4197 & 0.1016 & 1000 & 1000 & 100 \\
  \texttt{minesweeper} & rmsprop & 0.006163 & 9.3e-04 & 0.4410 & 0.2904 & 1000 & 2000 & 100 \\
  \texttt{roman-empire} & rmsprop & 0.003992 & 2.8e-05 & 0.5999 & 0.03923 & 1000 & 2000 & 100 \\
  \texttt{pokec} & adamw & 1.95e-04 & 1.27e-05 & 0.4399 & 0.3571 & 8000 & 1000 & 5 \\
  \texttt{ogbn-products} & adamw & 2.0e-04 & 6.0e-06 & 0.5 & 0.5 & 8000 & 1000 & 5 \\
  \texttt{ogbn-papers100M} & adamw & 1.8e-04 & 0 & 0.1555 & 0.1784 & 8000 & 100 & 5 \\
  \bottomrule
  \end{tabular}
  }
\end{table}

\begin{table}[ht]
  \centering
  \captionsetup{width=\textwidth}
  \caption{Architecture and routing hyperparameters. $K$ is the Chebyshev degree and $M$ is the number of filter experts. \emph{proj.} is the optional input-channel projection dimension applied to the Chebyshev basis; ``--'' means no projection. \emph{router} is either the Direct joint MLP router or the Response-aware two-stage router from Sec.~\ref{sec:method-routing}. $k$ is the sparse-routing top-$k$ value, with ``dense'' denoting dense softmax routing. $K_1$ is the optional Stage-1 candidate count for the response-aware two-stage router. $D$ is the response-key/query dimension.}
  \label{tab:hparams_arch}
  \scalebox{0.8}{
  \begin{tabular}[t]{lccccccccccc}
  \toprule
  Dataset & proj. & $K$ & $M$ & router & $k$ & $K_1$ & gate.h & gate.L & $D$ & head.h & head.L \\
  \midrule
  \texttt{coauthor-cs} & -- & 7 & 12 & Direct MLP & 2 & -- & 128 & 4 & -- & 512 & 3 \\
  \texttt{coauthor-physics} & -- & 7 & 6 & Direct MLP & 1 & -- & 128 & 2 & -- & 1024 & 4 \\
  \texttt{amazon-photo} & -- & 7 & 8 & Two-stage & 2 & 4 & 256 & 3 & 16 & 512 & 3 \\
  \texttt{amazon-computer} & 128 & 5 & 4 & Direct MLP & 1 & -- & 128 & 2 & -- & 256 & 4 \\
  \texttt{amazon-ratings} & -- & 14 & 10 & Direct MLP & dense & -- & 256 & 4 & -- & 1024 & 4 \\
  \texttt{tolokers} & 256 & 10 & 5 & Direct MLP & 2 & -- & 512 & 3 & -- & 512 & 2 \\
  \texttt{minesweeper} & 512 & 15 & 8 & Direct MLP & 1 & -- & 256 & 1 & -- & 128 & 2 \\
  \texttt{roman-empire} & 256 & 10 & 10 & Direct MLP & 1 & -- & 64 & 1 & -- & 128 & 3 \\
  \texttt{pokec} & 256 & 10 & 11 & Direct MLP & 1 & -- & 512 & 4 & -- & 1024 & 3 \\
  \texttt{ogbn-products} & -- & 10 & 10 & Two-stage & 2 & 2 & 256 & 3 & 32 & 512 & 3 \\
  \texttt{ogbn-papers100M} & 256 & 12 & 9 & Direct MLP & dense & -- & 768 & 4 & -- & 1024 & 3 \\
  \bottomrule
  \end{tabular}
  }
\end{table}

\begin{table}[ht]
  \centering
  \captionsetup{width=\textwidth}
  \caption{Auxiliary-loss coefficients and router-side structural encodings. $\lambda_{\mathrm{div}}$, $\lambda_{\mathrm{sm}}$, $\lambda_{\mathrm{imp}}$, $\lambda_{\mathrm{load}}$, and $\lambda_z$ correspond to the diversity, smoothness, importance, load, and router $z$-loss terms in Eq.~\eqref{eq:filtermoe-overall-objective}. A zero coefficient means the corresponding auxiliary term is inactive for that selected configuration. RWSE entries are written as hops/probes/proj.d; ``--'' means that random-walk structural encodings are not used by the router.}
  \label{tab:hparams_loss_rwse}
  \scalebox{0.90}{
  \begin{tabular}[t]{lcccccc}
  \toprule
  Dataset & $\lambda_{\mathrm{div}}$ & $\lambda_{\mathrm{sm}}$ & $\lambda_{\mathrm{imp}}$ & $\lambda_{\mathrm{load}}$ & $\lambda_z$ & RWSE \\
  \midrule
  \texttt{coauthor-cs} & 5.0e-04 & 0.05 & 0.05 & 0.2 & 0.01 & -- \\
  \texttt{coauthor-physics} & 5.0e-04 & 0.15 & 0 & 0.2 & 0.02 & 8/16/32 \\
  \texttt{amazon-photo} & 5.0e-04 & 0.15 & 0.4 & 0.2 & 0.002 & -- \\
  \texttt{amazon-computer} & 0 & 0.1 & 0 & 0.1 & 0 & 8/16/64 \\
  \texttt{amazon-ratings} & 0.0015 & 0.12 & 0.1 & 0.1 & 0.015 & -- \\
  \texttt{tolokers} & 0.006812 & 0.1392 & 0.3974 & 0.3811 & 0.03 & -- \\
  \texttt{minesweeper} & 0.001 & 0.2 & 0 & 0 & 0 & -- \\
  \texttt{roman-empire} & 1.0e-04 & 0.1 & 0 & 0 & 0 & -- \\
  \texttt{pokec} & 1.2e-05 & 0.2406 & 0.1170 & 0.1436 & 1.5e-03 & 8/16/64 \\
  \texttt{ogbn-products} & 0.001 & 0.2 & 0.1 & 0.2 & 0 & 8/16/32 \\
  \texttt{ogbn-papers100M} & 0.001751 & 0.1266 & 0.1 & 0.1 & 0.02703 & 32/64/128 \\
  \bottomrule
  \end{tabular}
  }
\end{table}

\section{Baseline Hyperparameter Search Spaces}
\label{app:baseline_hparams}

We use the accuracy numbers reported in the original baseline papers, and the benchmarking study of \citep{platonov2023critical}, as well as the ogb benchmark \citep{hu2020ogb} leaderboard, if the training protocol math our settings. For all other cases, this section summarizes the hyperparameter searches used for the baseline rows in
Tables~\ref{tab:main_homo}, \ref{tab:main_hetero}, and \ref{tab:main_large}. For each tuned
baseline, we run an Optuna TPE search on the validation split, select the best validation
configuration, and report the mean and standard deviation over $10$ random seeds. Unless otherwise
specified, each search uses $50$ trials per dataset. The same protocol is used for the
Chebyshev-operator variants of SIGN, HOGA, and GAMLP in Appendix~\ref{app:cheb_kernel_ablation}:
these variants are tuned independently after replacing the original random-walk-based diffusion
operator or normalized-adjacency preprocessing with the Chebyshev basis used by \modelname{}.
Thus, the Chebyshev-operator ablation changes the preprocessing operator but does not reuse the
hyperparameters selected for the original-operator run.

\paragraph{PP-GNN baselines.}
Table~\ref{tab:baseline_hparams_ppgnn} reports the search space for the PP-GNN baselines most
directly compared with \modelname{}. SIGN, HOGA, and GAMLP are evaluated with their original
preprocessing operators in the main tables and with the matched Chebyshev operator in
Appendix~\ref{app:cheb_kernel_ablation}. In both cases, the downstream hop-aggregation head is
tuned over the same ranges. GAMLP is evaluated without its optional label-reuse module, so the
reported numbers isolate the hop-feature aggregator rather than benefiting from an additional label
feedback loop.

\begin{table}[ht]
  \centering
  \captionsetup{width=\textwidth}
  \caption{Search ranges for PP-GNN baselines and ChebNetII. Continuous ranges are sampled uniformly unless marked log-uniform. Model-specific rows extend or override the common rows above.}
  \label{tab:baseline_hparams_ppgnn}
  \scalebox{0.78}{
  \begin{tabular}[t]{llp{7.5cm}}
  \toprule
  Block & Hyperparameter & Search range \\
  \midrule
  \multirow{8}{*}{\shortstack[l]{Common\\(SIGN, HOGA, GAMLP)}}
   & \texttt{lr} & $[10^{-4},\,5\times 10^{-2}]$ (log-uniform) \\
   & \texttt{weight\_decay} & $[10^{-7},\,5\times 10^{-3}]$ (log-uniform) \\
   & \texttt{hidden\_channels} & $\{128, 192, 256, 320, 384, 448, 512, 640\}$ \\
   & \texttt{dropout} & $[0.0,\,0.6]$ \\
   & \texttt{input\_dropout} & $[0.0,\,0.5]$ \\
   & \texttt{mlp\_hidden} & $\{128, 256, 320, 384, 512, 640\}$ \\
   & \texttt{mlp\_dropout} & $[0.0,\,0.6]$ \\
   & \texttt{mlplayers} & integer $[1,\,4]$ \\
  \midrule
  \multirow{4}{*}{SIGN}
   & \texttt{hidden\_channels} & $\{128, 256, 384, 512, 640, 768\}$ (overrides common) \\
   & \texttt{dropout} & $[0.0,\,0.7]$ (overrides common) \\
   & \texttt{num\_layers} & integer $[2,\,5]$ \\
   & preprocessing operator & original random-walk / normalized-adjacency operator, or matched Chebyshev operator in Appendix~\ref{app:cheb_kernel_ablation} \\
  \midrule
  \multirow{5}{*}{HOGA}
   & \texttt{num\_layers} & integer $[1,\,4]$ \\
   & \texttt{num\_heads} & $\{1,2,4,8\}$; the channel-head variant sets heads equal to the hidden dimension \\
   & \texttt{attn\_dropout} & $[0.0,\,0.5]$ \\
   & \texttt{use\_post\_res} & $\{0,1\}$ \\
   & preprocessing operator & original random-walk / normalized-adjacency operator, or matched Chebyshev operator in Appendix~\ref{app:cheb_kernel_ablation} \\
  \midrule
  \multirow{9}{*}{GAMLP}
   & \texttt{gamlp\_alpha} & $[0.1,\,0.9]$ \\
   & \texttt{gamlp\_n\_layers\_1} & integer $[2,\,5]$ \\
   & \texttt{gamlp\_n\_layers\_2} & integer $[2,\,5]$ \\
   & \texttt{gamlp\_input\_drop} & $[0.0,\,0.6]$ \\
   & \texttt{gamlp\_att\_drop} & $[0.0,\,0.6]$ \\
   & \texttt{gamlp\_act} & $\{\texttt{relu},\texttt{leaky\_relu},\texttt{sigmoid}\}$ \\
   & \texttt{gamlp\_pre\_process}, \texttt{gamlp\_residual} & $\{\textsc{False},\textsc{True}\}$ \\
   & \texttt{gamlp\_pre\_dropout}, \texttt{gamlp\_bns} & $\{\textsc{False},\textsc{True}\}$ \\
   & preprocessing operator & original random-walk / normalized-adjacency operator, or matched Chebyshev operator in Appendix~\ref{app:cheb_kernel_ablation} \\
  \midrule
  \multirow{8}{*}{ChebNetII}
   & \texttt{lr} & $[10^{-4},\,5\times 10^{-2}]$ (log-uniform) \\
   & \texttt{weight\_decay} & $[10^{-7},\,5\times 10^{-3}]$ (log-uniform) \\
   & \texttt{hidden\_channels} & $\{64,128,192,256,320,384,448,512,640,768\}$ \\
   & \texttt{num\_layers} & integer $[2,\,4]$ \\
   & \texttt{dropout} & $[0.0,\,0.6]$ \\
   & \texttt{chebnetii\_K} & $\{5,10,15\}$ \\
   & \texttt{chebnetii\_dprate} & $\{0.0,0.3,0.5\}$ \\
   & \texttt{chebnetii\_coeff\_activation} / \texttt{init\_cos\_sq} & $\{\texttt{none},\texttt{relu}\}$ / $\{0,1\}$ \\
  \bottomrule
  \end{tabular}
  }
\end{table}

\paragraph{Adaptive spectral, spatial, and graph-MoE baselines.}
For the remaining baselines, we use a shared optimization search space together with compact
model-specific choices. Table~\ref{tab:baseline_hparams_common} lists the common dimensions, and
Table~\ref{tab:baseline_hparams_specific} lists additional method-specific dimensions. These ranges
cover the adaptive spectral/spatial baselines ACMGCN, JacobiConv, ADC, DSF, NFGNN, AGDN-HA,
AGDN-HC, the graph-MoE baselines GMoE, Mowst, and NodeMoE, and LINKX where applicable. The
common rows apply unless a method-specific row explicitly overrides them.

\begin{table}[ht]
\centering
\captionsetup{width=\textwidth}
\caption{Shared Optuna search dimensions for adaptive spectral, spatial, graph-MoE, and LINKX baselines. Method-specific overrides are listed in Table~\ref{tab:baseline_hparams_specific}.}
\label{tab:baseline_hparams_common}
\begin{tabular}{ll}
\toprule
Hyperparameter & Search values \\
\midrule
\texttt{lr}              & $\{1\!\times\!10^{-4},\ 5\!\times\!10^{-4},\ 10^{-3},\ 5\!\times\!10^{-3},\ 10^{-2}\}$ \\
\texttt{weight\_decay}   & $\{0,\ 5\!\times\!10^{-5},\ 5\!\times\!10^{-4},\ 5\!\times\!10^{-3}\}$ \\
\texttt{dropout}         & $\{0.0,\ 0.2,\ 0.3,\ 0.5,\ 0.7\}$ \\
\texttt{hidden\_channels}& $\{64,\ 128,\ 256\}$ \\
\texttt{num\_layers}     & $\{2,\ 3\}$ \\
\bottomrule
\end{tabular}
\end{table}

\begingroup
\small
\setlength{\tabcolsep}{4pt}
\begin{longtable}{llp{8.2cm}}
\caption{Method-specific Optuna search dimensions for the non-PP-GNN baselines. The common rows in Table~\ref{tab:baseline_hparams_common} are added to each method unless explicitly overridden.}\label{tab:baseline_hparams_specific}\\
\toprule
Method & Hyperparameter & Search values \\
\midrule
\endfirsthead
\toprule
Method & Hyperparameter & Search values \\
\midrule
\endhead
\bottomrule
\endfoot

\multirow{3}{*}{ACMGCN}
 & \texttt{method} & $\{\texttt{acmgcn},\texttt{acmgcnp},\texttt{acmgcnpp}\}$ \\
 & \texttt{variant} & $\{0,1\}$ \\
 & \texttt{structure\_info} & $\{0,1\}$ \\
\midrule

\multirow{7}{*}{JacobiConv}
 & \texttt{jacobi\_depth} & $\{5,10,15\}$ \\
 & \texttt{jacobi\_alpha} & $\{0.5,1.0,2.0\}$ \\
 & \texttt{jacobi\_a} & $\{-0.5,0.0,0.5,1.0,1.5,2.0\}$ \\
 & \texttt{jacobi\_b} & $\{-0.5,0.0,0.5,1.0,1.5,2.0\}$ \\
 & \texttt{jacobi\_aggr} & $\{\texttt{gcn},\texttt{mean}\}$ \\
 & \texttt{jacobi\_dpb} & $\{0.0,0.3,0.5\}$ \\
 & \texttt{jacobi\_dpt} & $\{0.0,0.3,0.5\}$ \\
\midrule

\multirow{3}{*}{ADC}
 & \texttt{adc\_init\_t} & $\{0.5,1.0,2.0,3.0\}$ \\
 & \texttt{adc\_step} & $\{5,10,15\}$ \\
 & \texttt{adc\_dense\_t} & $\{0,1\}$ \\
\midrule

\multirow{7}{*}{DSF}
 & \texttt{dsf\_K} & $\{5,10,15\}$ \\
 & \texttt{dsf\_alpha} & $\{0.05,0.1,0.3,0.5\}$ \\
 & \texttt{dsf\_dprate} & $\{0.0,0.3,0.5\}$ \\
 & \texttt{dsf\_pe\_hid\_dim} & $\{16,32,64\}$ \\
 & \texttt{dsf\_pe\_alpha} & $\{0.3,0.5,0.7\}$ \\
 & \texttt{dsf\_pe\_dropout} & $\{0.0,0.3\}$ \\
 & \texttt{dsf\_init} & $\{\texttt{PPR},\texttt{Random}\}$ \\
\midrule

\multirow{5}{*}{NFGNN}
 & \texttt{nfgnn\_K} & $\{5,10\}$ \\
 & \texttt{nfgnn\_alpha} & $\{0.05,0.1,0.3,0.5\}$ \\
 & \texttt{nfgnn\_rank} & $\{1,2,4\}$ \\
 & \texttt{nfgnn\_dprate} & $\{0.0,0.3,0.5\}$ \\
 & \texttt{nfgnn\_init} & $\{\texttt{PPR},\texttt{Random},\texttt{Fix}\}$ \\
\midrule

\multirow{10}{*}{AGDN-HA / AGDN-HC}
 & \texttt{agdn\_num\_heads} & $\{1,2,3,4\}$ \\
 & \texttt{agdn\_K} & $\{2,3,5\}$ \\
 & \texttt{agdn\_input\_dropout} & $\{0.0,0.1,0.3\}$ \\
 & \texttt{agdn\_attn\_drop} & $\{0.0,0.1,0.3\}$ \\
 & \texttt{agdn\_edge\_drop} & $\{0.0,0.1,0.3\}$ \\
 & \texttt{agdn\_transition\_matrix} & $\{\texttt{gat},\texttt{gat\_adj},\texttt{gat\_sym}\}$ \\
 & \texttt{agdn\_residual} & $\{0,1\}$ \\
 & \texttt{agdn\_position\_emb} & $\{0,1\}$ \\
 & \texttt{agdn\_HA\_activation} (HA only) & $\{\texttt{leakyrelu},\texttt{sigmoid},\texttt{relu}\}$ \\
 & weight style & fixed to \texttt{HA} or \texttt{HC} for the corresponding row \\
\midrule

\multirow{5}{*}{GMoE}
 & \texttt{num\_layers} & $\{3,4,5\}$ (overrides common) \\
 & \texttt{gmoe\_backbone} & $\{\texttt{gcn},\texttt{sage}\}$ \\
 & \texttt{gmoe\_num\_experts} & $\{2,4,8\}$ \\
 & \texttt{gmoe\_top\_k} & $\{1,2\}$ \\
 & \texttt{gmoe\_coef} & $\{0,10^{-3},10^{-2},10^{-1}\}$ \\
\midrule

\multirow{4}{*}{Mowst}
 & \texttt{mowst\_strong\_backbone} & $\{\texttt{gcn},\texttt{sage}\}$ \\
 & \texttt{mowst\_confidence} & $\{\texttt{learnable},\texttt{variance}\}$ \\
 & \texttt{mowst\_alpha} & $\{0.5,1.0,2.0\}$ \\
 & \texttt{mowst\_batch\_norm} & $\{0,1\}$ \\
\midrule

\multirow{8}{*}{NodeMoE}
 & \texttt{nodemoe\_num\_experts} & $\{2,3\}$ \\
 & \texttt{nodemoe\_gate\_hidden} & $\{32,64\}$ \\
 & \texttt{nodemoe\_gate\_layers} & $\{2,3\}$ \\
 & \texttt{nodemoe\_gate\_dropout} & $\{0.0,0.5,0.8\}$ \\
 & \texttt{nodemoe\_dprate} & $\{0.0,0.5\}$ \\
 & \texttt{nodemoe\_topk} & $\{0\text{ (soft)},1\}$ \\
 & \texttt{nodemoe\_smooth\_w} & $\{0,0.01,0.1,1.0\}$ \\
 & \texttt{nodemoe\_lb\_w} & $\{0,10^{-3},10^{-2},10^{-1}\}$ \\
\midrule

\multirow{4}{*}{LINKX}
 & \texttt{linkx\_init\_layers\_A} & $\{1,2\}$ \\
 & \texttt{linkx\_init\_layers\_X} & $\{1,2\}$ \\
 & \texttt{linkx\_inner\_activation} & $\{0,1\}$ \\
 & \texttt{linkx\_inner\_dropout} & $\{0,1\}$ \\

\end{longtable}
\endgroup

\paragraph{LINKX on \texttt{ogbn-papers100M}.}
The \texttt{ogbn-papers100M} run for LINKX uses a reduced search space to fit the full graph within
GPU memory. We search over the final MLP depth, dropout, learning rate, weight decay, and the
LINKX initialization flags, while fixing the largest budget-sensitive values: hidden width $16$, $30$
epochs, patience $8$, batch size $30000$, and evaluation batch size $20000$.

\section{Limitations and Future Work}
\label{app:limitations}

Our study focuses on the setting for which pre-propagation GNNs are most directly designed: node classification on fixed, homogeneous graphs. Here \emph{homogeneous} means that the graph has a single node/edge type and can be represented by one propagation operator. This controlled scope allows us to isolate the effect of joint node--channel filter routing, but it also leaves several extensions open.

\paragraph{Task scope.}
All experiments in this paper are node-classification benchmarks. Extending \modelname{} to link prediction would require combining two endpoint representations, or learning pair-specific routing/decoding mechanisms, while preserving the preprocessing-only training path. Extending it to graph-level prediction would require a graph-level readout over routed node representations and a batching strategy for multiple graphs with different sizes and spectral distributions. These task settings may require different regularization and routing choices from those used in node classification.

\paragraph{Graph type.}
The current formulation assumes a homogeneous graph and constructs a single Chebyshev basis from one graph operator. Heterogeneous graphs introduce node types, edge types, and relation-specific semantics, for which a single propagation operator may be insufficient. A natural extension is to build relation-aware or meta-path-aware basis tensors and route over type-conditioned filter experts. We leave such heterogeneous-graph extensions to future work.

\paragraph{Temporal setting.}
Our method also assumes a static graph during preprocessing and training. For dynamic graphs, changes in edges, node features, or node sets can invalidate the cached Chebyshev basis. Recomputing the full basis after every update would weaken the scalability benefit of pre-propagation. Future work could combine FilterMoE-style routing with incremental basis updates, temporal filter experts, or online routers that adapt to evolving graph structure.

Overall, these limitations reflect the current experimental scope rather than restrictions inherent to node--channel filter routing. Exploring link prediction, graph-level prediction, heterogeneous graphs, and dynamic graphs are promising directions for extending the pre-propagation filter-routing framework.

\section{Potential Societal Impacts}
\label{app:societal_impacts}

This work is a methodological study of scalable pre-propagation GNNs, evaluated on node-classification benchmarks. It does not introduce a deployed system or a new dataset of sensitive personal information. Nevertheless, graph-learning methods are often applied to social, web, biological, financial, and recommendation networks, so improvements in scalability and accuracy can have both positive and negative downstream effects.

\paragraph{Potential positive impacts.}
By preserving the preprocessing-only training path of \PPGs{}, \modelname{} can make graph learning more practical on large relational datasets. More efficient training may reduce the computational cost of model selection, lower the barrier for researchers with limited hardware, and decrease the energy required for repeated experimentation. The method may also be useful in beneficial graph-analysis settings, such as scientific citation analysis, biological network modeling, infrastructure monitoring, fraud detection, and information retrieval, where scalable node representations can help identify relevant entities or patterns. In addition, replacing dataset-specific manual hop-aggregator selection with learned node--channel filter routing may improve reproducibility by reducing the amount of hand-tuning needed to obtain strong PP-GNN performance.

\paragraph{Potential negative impacts.}
The same scalability benefits can also make it easier to apply graph learning to sensitive relational data. In social or behavioral networks, node attributes and graph structure may encode demographic, socioeconomic, or other sensitive information. A more accurate or scalable node-classification model could therefore amplify biased labels, enable intrusive profiling, or support surveillance and targeted manipulation if used without appropriate governance. Because PP-GNNs cache propagated features, the cached basis may also reveal information about local neighborhoods; deployments involving private graphs should therefore consider data minimization, access control, privacy-preserving preprocessing, and secure storage of cached features.

\paragraph{Mitigation and responsible use.}
We recommend treating \modelname{} as a general-purpose graph representation method rather than a deployment-ready decision system. Responsible use should include documentation of graph construction choices, evaluation across relevant subpopulations, comparison to simpler baselines, monitoring under distribution shift, and privacy review when graph edges or node features are sensitive. These safeguards are especially important because scalability improvements can broaden the set of datasets and organizations for which large-scale graph learning becomes feasible.

\section{Existing Assets, Licenses, and Terms of Use}
\label{app:asset_licenses}

This paper uses existing assets only for benchmarking: public node-classification datasets, standard data loaders, and baseline model implementations or reimplementations. We cite the original papers in the main text and appendices. We do not redistribute raw third-party datasets or third-party baseline repositories with this submission. When an official dataset or repository page does not expose an explicit license, we mark the license as \emph{not specified} rather than assuming a permissive license, and use the asset only as a cited research benchmark or through our own implementation.

\paragraph{Datasets.}
Table~\ref{tab:asset_license_datasets} summarizes the datasets used in the experiments. The OGB pages explicitly list licenses for \texttt{ogbn-products} and \texttt{ogbn-papers100M}. The heterophilous datasets are distributed through the official \texttt{yandex-research/heterophilous-graphs} repository, which is MIT licensed; the repository also documents the underlying sources for \texttt{roman-empire}, \texttt{amazon-ratings}, \texttt{minesweeper}, and \texttt{tolokers}. For the PyG Amazon/Coauthor benchmarks and the SNAP \texttt{pokec} benchmark, the public dataset pages provide source/citation information, but we did not find a dataset-specific license on the official pages used in this study.

\begin{table*}[t]
  \centering
  \captionsetup{width=0.96\textwidth}
  \caption{Dataset assets used in this paper. ``Not specified'' means that we found source/citation information on the official page used for the benchmark, but did not find an explicit dataset-specific license. We do not redistribute raw third-party datasets.}
  \label{tab:asset_license_datasets}
  \scriptsize
  \setlength{\tabcolsep}{3.5pt}
  \renewcommand{\arraystretch}{1.15}
  \begin{tabularx}{\textwidth}{>{\raggedright\arraybackslash}p{2.9cm}>{\raggedright\arraybackslash}p{4.0cm}>{\raggedright\arraybackslash}X}
  \toprule
  Dataset(s) & Source / owner & License or terms noted \\
  \midrule
  \texttt{coauthor-cs}, \texttt{coauthor-physics} & PyTorch Geometric \texttt{Coauthor} loader; processed benchmark from \citet{shchur1811pitfalls} & PyG code/loaders are MIT licensed, and the processed benchmark repository from \citet{shchur1811pitfalls} is MIT licensed. A separate raw-dataset license is not specified on the PyG dataset page; the original benchmark source is credited and the raw data are not redistributed. \\
  \texttt{amazon-photo}, \texttt{amazon-computer} & PyTorch Geometric \texttt{Amazon} loader; processed benchmark from \citet{shchur1811pitfalls} & PyG code/loaders are MIT licensed, and the processed benchmark repository from \citet{shchur1811pitfalls} is MIT licensed. A separate raw-dataset license is not specified on the PyG dataset page; the original benchmark source is credited and the raw data are not redistributed. \\
  \texttt{roman-empire}, \texttt{amazon-ratings}, \texttt{minesweeper}, \texttt{tolokers} & GitHub: \texttt{yandex-research/}\linebreak[1]\texttt{heterophilous-graphs} & MIT license for the repository. The repository attributes the underlying data sources: English Wikipedia for \texttt{roman-empire}, SNAP for \texttt{amazon-ratings}, a synthetic generator for \texttt{minesweeper}, and Toloka data for \texttt{tolokers}. \\
  \texttt{pokec} & SNAP Pokec dataset page & The SNAP Pokec page provides dataset statistics and citation/source information. The SNAP software/site license page states a BSD license for SNAP, but we did not find an explicit dataset-specific license on the Pokec dataset page; the dataset is used only for research evaluation and is not redistributed. \\
  \texttt{ogbn-products} & Open Graph Benchmark node-property-prediction page & OGB lists the license as the Amazon license for \texttt{ogbn-products}. \\
  \texttt{ogbn-papers100M} & Open Graph Benchmark node-property-prediction page & OGB lists the license as ODC-BY for \texttt{ogbn-papers100M}. \\
  \bottomrule
  \end{tabularx}
\end{table*}

\paragraph{Baseline implementations and libraries.}
Table~\ref{tab:asset_license_baselines} lists the code assets used to implement, reproduce, or compare against baselines. For baselines whose official repositories provide explicit licenses, we list the license name. For baselines without an explicit repository license, we either use our own implementation guided by the paper or use the public code only for reference and do not redistribute it.

\begin{table*}[t]
  \centering
  \captionsetup{width=0.96\textwidth}
  \caption{Baseline and library assets. For repositories with no visible license, we report ``not specified'' and do not redistribute their code.}
  \label{tab:asset_license_baselines}
  \scriptsize
  \setlength{\tabcolsep}{3.5pt}
  \renewcommand{\arraystretch}{1.15}
  \begin{tabularx}{\textwidth}{>{\raggedright\arraybackslash}p{3.0cm}>{\raggedright\arraybackslash}p{4.1cm}>{\raggedright\arraybackslash}X}
  \toprule
  Asset / baseline & Source & License or terms noted \\
  \midrule
  PyTorch Geometric utilities; GCN, GraphSAGE, and GAT layers/loaders & GitHub: \texttt{pyg-team/}\linebreak[1]\texttt{pytorch\_geometric} & MIT license. \\
  SIGN & GitHub: \texttt{twitter-research/}\linebreak[1]\texttt{sign} & Apache-2.0 license. \\
  HOGA & GitHub: \texttt{cornell-zhang/}\linebreak[1]\texttt{HOGA} & BSD 3-Clause license. \\
  GAMLP & GitHub: \texttt{PKU-DAIR/}\linebreak[1]\texttt{GAMLP} & Public research repository; no explicit license found in this audit. We cite the original work and do not redistribute the repository. \\
  ChebNetII & GitHub: \texttt{ivam-he/}\linebreak[1]\texttt{ChebNetII} & Public research repository; no explicit license found in this audit. We cite the original work and do not redistribute the repository. \\
  GPRGNN & GitHub: \texttt{jianhao2016/}\linebreak[1]\texttt{GPRGNN} & Public research repository; no explicit license found in this audit. We cite the original work and do not redistribute the repository. \\
  ACMGCN & GitHub: \texttt{SitaoLuan/}\linebreak[1]\texttt{ACM-GNN} & MIT license. \\
  ADC & GitHub: \texttt{abcbdf/}\linebreak[1]\texttt{ADC} & MIT license. \\
  JacobiConv & GitHub: \texttt{GraphPKU/}\linebreak[1]\texttt{JacobiConv} & Public research repository; no explicit license found in this audit. We cite the original work and do not redistribute the repository. \\
  DSF & Original paper / public implementation when available & No explicit license identified in this audit. We cite the original work and use our own implementation or reported settings where needed. \\
  NFGNN & GitHub: \texttt{SsGood/}\linebreak[1]\texttt{NFGNN} & Public research repository; no explicit license found in this audit. We cite the original work and do not redistribute the repository. \\
  AGDN-HA, AGDN-HC & GitHub: \texttt{skepsun/}\linebreak[1]\texttt{Adaptive-Graph-}\linebreak[1]\texttt{Diffusion-Networks} & MIT license. \\
  GMoE & GitHub: \texttt{VITA-Group/}\linebreak[1]\texttt{Graph-Mixture-}\linebreak[1]\texttt{of-Experts} & MIT license. \\
  Mowst & GitHub: \texttt{facebookresearch/}\linebreak[1]\texttt{mowst-gnn} & MIT license. \\
  NodeMoE & Original paper / OpenReview record & No public repository license identified in this audit. We cite the original work and do not redistribute third-party code. \\
  LINKX, where used & GitHub: \texttt{CUAI/}\linebreak[1]\texttt{Non-Homophily-}\linebreak[1]\texttt{Large-Scale}; original LINKX paper & Used only as a cited baseline/reference where applicable; we do not redistribute third-party code or data from this source. \\
  \bottomrule
  \end{tabularx}
\end{table*}


\paragraph{Compliance practice.}
For permissively licensed code assets, any released implementation should preserve the corresponding copyright and license notices. For datasets or code with unspecified licenses, we avoid redistribution and provide only citations, experimental results, and implementation details sufficient for reproduction. If the code associated with this paper is released, we will include a third-party notices file summarizing the assets in Tables~\ref{tab:asset_license_datasets}--\ref{tab:asset_license_baselines}.

\end{document}